\RequirePackage{fix-cm}
\documentclass[smallextended]{svjour3}       % onecolumn (second format)
\smartqed  % flush right qed marks, e.g. at end of proof

\usepackage{enumitem}
\usepackage{graphicx}
\usepackage{amssymb}
\usepackage{tabularx}
\usepackage{xcolor}
\usepackage{fontenc}
\usepackage{amsmath}
\usepackage{colortbl}
\usepackage{multirow}
\usepackage{array}
\usepackage{pgfplots}
\usepackage{pgfplots,lipsum}
\usepackage{tikz}
\usetikzlibrary{patterns}
\usepackage{makecell}
\usepackage{hhline}
\usepackage{lipsum}
\usepackage{algorithm}
\usepackage{algpseudocode}
\usepackage{algcompatible}
\usepackage{pdfpages}
\usepackage{adjustbox}
\usepackage{setspace}
\usepackage{lscape}
\usepackage[sort&compress,numbers]{natbib}
\usepackage{lineno,hyperref}
\usepackage[margin=1in]{geometry}	%margine changes

\usepackage[normalem]{ulem}	% underline a paragraph

\algrenewcommand\algorithmicindent{2.0em}%

\algdef{SE}{MyFor}{EndMyFor}[1]{%
	\textbf{foreach} #1}{%
	\textbf{end foreach}}

\begin{document}

\title{Automatic Personality Prediction; an Enhanced Method Using Ensemble Modeling}


\author{Majid Ramezani$^1$	\and
        Mohammad-Reza Feizi-Derakhshi*$^{,1}$	\and
        Mohammad-Ali Balafar$^2$	\and
        Meysam Asgari-Chenaghlu$^1$	\and
        Ali-Reza Feizi-Derakhshi$^1$	\and
        Narjes Nikzad-Khasmakhi$^1$	\and
        Mehrdad Ranjbar-Khadivi$^{3,1}$	\and
        Zoleikha Jahanbakhsh-Nagadeh$^{4,1}$	\and
        Elnaz Zafarani-Moattar$^{5,1}$	\and
        Taymaz Rahkar-Farshi$^{6,1}$
}


\institute{
	\\
		* Corresponding author: Mohammad-Reza Feizi-Derakhshi (\email{mfeizi@tabrizu.ac.ir}; Tell-Fax: +984133393760)\\
		\\
		\\
		(1) \at 
              Computerized Intelligence Systems Laboratory, Department of Computer Engineering, University of Tabriz, Tabriz, Iran. \\
              %\email{m\_ramezani@tabrizu.ac.ir}           %  \\
           %\and
           %Mohammad-Reza Feizi-Derakhshi \at
              %Computerized Intelligence Systems Laboratory, Department of Computer Engineering, University of Tabriz, Tabriz, Iran. \\
              %\email{mfeizi@tabrizu.ac.ir}
           \and
           (2)	\at
              Department of Computer Engineering, University of Tabriz, Iran.	\\
              %\email{balafarila@tabrizu.ac.ir}
           \and
           %Meysam Asgari-Chenaghlu	\at
              %Computerized Intelligence Systems Laboratory, Department of Computer Engineering, University of Tabriz, Tabriz, Iran. \\
              %\email{m.asgari@tabrizu.ac.ir}
           %\and
           %Ali-Reza Feizi-Derakhshi	\at
             % Computerized Intelligence Systems Laboratory, Department of Computer Engineering, University of Tabriz, Tabriz, Iran. \\
              %\email{derakhshi96@ms.tabrizu.ac.ir}
           %\and
           %Narjes Nikzad-Khasmakhi	\at
            %  Computerized Intelligence Systems Laboratory, Department of Computer Engineering, University of Tabriz, Tabriz, Iran. \\
              %\email{n.nikzad@tabrizu.ac.ir}
           %\and
           (3)	\at
           	 Department of Computer Engineering, Shabestar Branch, Islamic Azad University, Shabestar, Iran.	\\
           	 %\email{mehrdad.khadivi@iaushab.ac.ir}
           \and
           (4)	\at
	           Department of Computer Engineering, Naghadeh Branch, Islamic Azad University, Naghadeh, Iran.	\\
	           %\email{zoleikha.jahanbakhsh@srbiau.ac.ir}
	       \and
	       (5)	\at
		       Department of Computer Engineering, Tabriz Branch, Islamic Azad University, Tabriz, Iran.	\\
		       %\email{e.zafarani@iaut.ac.ir}
		   \and
		   (6)	\at
			   Department of Software Engineering, Altinbas University, Istanbul, Turkey.	\\
			   \\
			   \\
			   %\email{taymaz.farshi@altinbas.edu.tr}
		\and	
		Majid Ramezani(\email{m\_ramezani@tabrizu.ac.ir}; ORCID: 0000-0003-0886-7023);\\
		Mohammad-Reza Feizi-Derakhshi (\email{mfeizi@tabrizu.ac.ir}; ORCID: 0000-0002-8548-976X);\\
		Mohammad-Ali Balafar(\email{balafarila@tabrizu.ac.ir}; ORCID: 0000-0001-5898-0871);\\
		Meysam Asgari-Chenaghlu(\email{m.asgari@tabrizu.ac.ir}; ORCID: 0000-0002-7892-9675);\\
		Ali-Reza Feizi-Derakhshi (\email{derakhshi96@ms.tabrizu.ac.ir}; ORCID: 0000-0003-3036-1651);\\
		Narjes Nikzad-Khasmakhi(\email{n.nikzad@tabrizu.ac.ir});\\
		Mehrdad Ranjbar-Khadivi(\email{mehrdad.khadivi@iaushab.ac.ir}; ORCID: 0000-0002-3441-8224);\\
		Zoleikha Jahanbakhsh-Nagadeh(\email{zoleikha.jahanbakhsh@srbiau.ac.ir});\\
		Elnaz Zafarani-Moattar(\email{e.zafarani@iaut.ac.ir});\\
		Taymaz Rahkar-Farshi(\email{taymaz.farshi@altinbas.edu.tr}; ORCID: 0000-0003-4070-1058)
		\\
		\\
		\\
		\\
		\\
		%\and
		%\textbf{Conflict of Interest}: Author A (Majid Ramezani) declares that he has no conflict of interest. Author B (Mohammad-Reza Feizi-Derakhshi) declares that he has no conflict of interest. Author C (Mohammad-Ali Balafar) declares that he has no conflict of interest. Author D (Meysam Asgari-Chenaghlu) declares that he has no conflict of interest. Author E (Ali-Reza Feizi-Derakhshi) declares that he has no conflict of interest. Author F (Narjes Nikzad-Khasmakhi) declares that she has no conflict of interest. Author G (Mehrdad Ranjbar-Khadivi) declares that he has no conflict of interest. Author H (Zoleikha Jahanbakhsh-Nagadeh) declares that she has no conflict of interest. Author I (Elnaz Zafarani-Moattar) declares that she has no conflict of interest. Author J (Taymaz Rahkar-Farshi) declares that he has no conflict of interest.
}


\maketitle

\begin{abstract}
Human personality is significantly represented by those words which he/she uses in his/her speech or writing. As a consequence of spreading the information infrastructures (specifically the Internet and social media), human communications have reformed notably from face to face communication. Generally, Automatic Personality Prediction (or Perception) (APP) is the automated forecasting of the personality  on different types of human generated/exchanged contents (like text, speech, image, video, etc.). The major objective of this study is to enhance the accuracy of APP from the text. To this end, we suggest five new APP methods including term frequency vector-based, ontology-based, enriched ontology-based, latent semantic analysis (LSA)-based, and deep learning-based (BiLSTM) methods. These methods as the base ones, contribute to each other to enhance the APP accuracy through ensemble modeling (stacking) based on a hierarchical attention network (HAN) as the meta-model. The results show that ensemble modeling enhances the accuracy of APP.
\end{abstract}

\section{Introduction} \label{intro}
\textit{Personality} is defined as the characteristic set of behaviors, cognitions, and emotional patterns \cite{corr2009cambridge} as well as thinking patterns \cite{kazdin2000encyclopedia}. Analyzing the personality of people, psychologists achieve these patterns and find what makes them, who they are. It serves a variety of purposes such as analyzing health, personality and mental disorders, work and academic successes, matrimonial stability, friendships, political psychology, human resource employment and management, marketing and customer behaviors,  etc. The\textit{ Automatic Personality Prediction} (or \textit{Perception}) (\textit{APP}) is the automatic prediction of the personality of individuals \cite{7899604}. Nowadays, people interact with each other through miscellaneous information infrastructures including social media (with different data types like text, image, voice and video and also different frameworks), image and video sharing networks, emails, short message services (SMS), etc. All of them, as a kind of human interactions, provide various presentations of people's personality; even better than the real world. Discovering people's behavioral, cognitive, emotional and thinking patterns, without facing them, would be interesting and may have different objectives. 

Recent trends in APP have led to a proliferation of studies through analyzing different data types, in particular analyzing speech \cite{7344614,10.1007/978-3-319-14445-0_16, Jothilakshmi2017}, image \cite{10.1007/978-3-030-00021-9_54, Celli:2014:API:2647868.2654977, 10.1007/978-3-319-27671-7_71}, video \cite{10.1007/978-3-319-93034-3_51, Gucluturk_2017_ICCV, 10.1007/978-3-319-49409-8_25}, , social media contents and activities \cite{10.1007/978-981-13-9187-3_65, chen2016user, 8494744}, emails \cite{10.1007/978-3-642-38844-6_29}, handwriting \cite{8769221, wijaya2018personality}, touch screen-based interaction \cite{8463375}, signature  \cite{8122145}, nonverbal behaviors \cite{7163171}, mobile short message services (SMS) \cite{10.1007/978-3-319-21206-7_44}, and so on. Along with this growth in APP, however there is an increasing concern over the accuracy of the predictions. The major objective of present study was to suggest an enhanced method to improve the accuracy of APP, using ensemble modeling. Thereby, the key research question of this study was whether or not applying ensemble modeling would help to improve the accuracy of APP? To do so we have suggested five new APP methods, including \textit{term frequency vector-based}, \textit{ontology-based}, \textit{enriched ontology-based}, \textit{latent semantic analysis-based}, and \textit{deep learning-based} methods. All of them, as the independent base methods will cooperate together to reduce the generalization error of the prediction, through ensemble modeling \cite{kotu2018data}. Actually,  ensemble modeling is a process where multiple diverse base models are used to predict an outcome \cite{kotu2018data} rather than any of the constituent models alone. So the hypothesis that will be tested is that the prediction error decreases when the ensemble method is used. 

To date, several personality trait models have been introduced. Allport’s trait theory \cite{allport1937personality}, Cattell’s 16 Factor Model \cite{cattell1970handbook}, Eysenck’s Giant Three \cite{eysenek1950dimensions},  Myers–Briggs Type Indicator (MBTI)  \cite{briggs1976myers}, and the Big Five Model (Five Factor Model) amongst others are considered as some of the most important models \cite{mccrae1992introduction}, in which the last one is the most widely accepted trait model, presently \cite{Matz2016}. The Big Five personality traits foundation, has borrowed from lexical hypothesis \cite{goldberg1993structure}. It basically states that people encode in their everyday language all those essential personality characteristics and individual differences \cite{Uher2013}. Furthermore, the Big Five model appears to be nearly universally held, independent of cultures \cite{mccrae2005universal}. It basically describes the personality traits in five broad categories: \textit{openness}, \textit{conscientiousness}, \textit{extroversion}, \textit{agreeableness}, and \textit{neuroticism} \cite{john1999big}. Usually, they are called by the acronym \textit{OCEAN} or \textit{CANOE}. Each of the five personality factors or traits represents a range between two extremes; to be specific, extroversion represents a continuum, between extreme extroversion and extreme introversion \cite{cherry2019big}. \hyperref[tab: Table1]{Table~\ref*{tab: Table1}}, illustrates some of the main characteristics of each of the five traits.

It is worthwhile noting that, due to the fact that the Big Five model is founded based on lexical hypothesis, lexical methods are selected to carry out the personality prediction. This is the reason why we have selected lexical methods like term frequency-based, ontology-based and LSA-based methods for APP. Then these methods are diversified by a deep learning-based method to be applied in ensemble modeling.

The present study makes two noteworthy contributions to APP from text as follows: \textit{i)} developing five new methods to APP, \textit{ii)} enhancing the personality prediction accuracy, by establishing an ensemble model. 

The rest of this study is organized in the following manner: the second section gives a brief literature review of APP. In the third section, five proposed new methods for APP as well as ensemble modeling architecture are described. The experimental results are presented and discussed respectively in the fourth and fifth section. Some conclusions are drawn in the final section, and the areas for further research are identified, also.

\begin{table}[htbp]
	\scriptsize
	%\tiny
	%\fontsize{7}{8}
	%body of the table
	%\resizebox{\textwidth}{!}
	{
		\begin{center}
			\caption{An overview of five factor personality traits (the Big Five model)}
			\label{tab: Table1}
			\fontsize{7}{8}
			\begin{adjustbox}{max width=\textwidth}
				\begin{tabular}{@{}p{5.5cm}@{}p{3.2cm}@{}p{5.5cm}@{}} % <-- Alignments: 1st column left, 2nd middle and 3rd right, with vertical lines in between
					
					\hline
					\hline
					\cellcolor{gray!15}	\textbf{Description of \textit{LOW} values \textcolor{gray} {-}} 	\textcolor{gray} {$\infty$ $\longleftarrow$}  	&	 \cellcolor{gray!15}	\textbf{Personality trait} \centering	&  		\cellcolor{gray!15}	\textcolor{gray} {$\longrightarrow$ +$\infty$}	 \textbf{ Description of \textit{HIGH} values}\\
					\hline
					\hline
					%--------   Openness ---------
					& \textbf{Openness (O)} \centering & 	
					\\
					\hline
					
					\begin{tabular}{l}
						\tiny		%  To minimize the font size
						\begin{minipage}[t]{0.35\columnwidth}		% In order to Wrap text and evoide line exceedings to next cells
							•	Dislikes changes \\		
							•	Does not enjoy new things \\
							•	Conventional\\
							•	Resists new ideas\\	
							•	Prefers familiarity\\
							•	Not very imaginative\\
							•	Has trouble with abstract or theoretical concepts\\
							•	Skeptical\\
							•	Traditional in thinking\\
							•	Consistent and cautious
						\end{minipage}				% In order to Wrap text and evoide line exceedings to next cells
					\end{tabular}
					
					&& 
					\begin{tabular}{l}
						\tiny		%  To minimize the font size
						\begin{minipage}[t]{0.35\columnwidth}		% In order to Wrap text and evoide line exceedings to next cells
							•	Very creative\\	
							•	Clever, insightful, daring, and varied interests\\
							•	Embraces trying new things or visiting new places\\
							•	Unconventional\\
							•	Focused on tackling new challenges\\
							•	Intellectually curious\\
							•	Inventive\\
							•	Happy to think about abstract concepts\\
							•	Enjoys the art\\
							•	Eager to meet new people
						\end{minipage}
					\end{tabular}
					
					\\
					\hline

					%--------   Conscientiousness ---------
					& 	\textbf{Conscientiousness (C)} 	\centering	 & 
					
					\\
					\hline
					
					\begin{tabular}{l}
						\tiny		%  To minimize the font size
						\begin{minipage}[t]{0.35\columnwidth}		% In order to Wrap text and evoide line exceedings to next cells
							•	Easy going and careless\\
							•	Messy and less detailed-oriented\\
							•	Dislikes structure and schedule\\
							•	Fails to return things or put them back, where they belong\\
							•	Procrastinates on important tasks and rarely completes them on time\\
							•	Fails to stick to a schedule\\
							•	Is always late when meeting others
						\end{minipage}				% In order to Wrap text and evoide line exceedings to next cells
					\end{tabular}
					
					& & 
					
					\begin{tabular}{l}
						\tiny		%  To minimize the font size
						\begin{minipage}[t]{0.35\columnwidth}		% In order to Wrap text and evoide line exceedings to next cells
							•	Competent and efficient\\
							•	Goal- and detail-oriented\\
							•	Well organized, self-discipline and dutiful\\
							•	Spends time preparing\\
							•	Predictable and deliberate\\
							•	Finishes important tasks on time\\
							•	Does not give in to impulses\\
							•	Enjoys adhering to a schedule\\
							•	Is on time when meeting others\\
							•	Works hardly\\ 
							•	Reliable and resourceful\\
							•	Persevered
						\end{minipage}				% In order to Wrap text and evoide line exceedings to next cells
					\end{tabular}
					
					\\
					\hline

					%--------   Extraversion (E) ---------
					&	 \textbf{Extroversion (E)} 		\centering	 & 
					\\
					\hline
					
					\begin{tabular}{l}
						\tiny		%  To minimize the font size
						\begin{minipage}[t]{0.35\columnwidth}		% In order to Wrap text and evoide line exceedings to next cells
							•	Introspective\\
							•	Solitary and reserved\\
							•	Dislikes being at the center of attentions\\
							•	Feels exhausted when having to socialize a lot\\
							•	Finds it difficult to start conversations\\
							•	Dislikes making small talks\\
							•	Carefully thinks things before speaking\\
							•	Thoughtful
						\end{minipage}				% In order to Wrap text and evoide line exceedings to next cells
					\end{tabular}
					
					& & 
					
					\begin{tabular}{l}
						\tiny		%  To minimize the font size
						\begin{minipage}[t]{0.35\columnwidth}		% In order to Wrap text and evoide line exceedings to next cells
							•	Outgoing and energetic\\
							•	Assertive and talkative\\
							•	Able to be articulate\\
							•	Enjoys being the center of attentions\\
							•	Likes to start conversations\\
							•	Enjoys being with others and meeting new people\\
							•	Tendency to be affectionate\\
							•	Finds it easy to make new friends\\
							•	Has a wide social circle of friends and acquaintances\\
							•	Says things before thinking about them\\
							•	Feels organized when around other people\\
							•	Social confidence
						\end{minipage}				% In order to Wrap text and evoide line exceedings to next cells
						
					\end{tabular}
					
					\\
					\hline

					%--------   Agreeableness (A) ---------
					&	 \textbf{Agreeableness (A)}  \centering		& 
					\\
					\hline
					
					\begin{tabular}{l}
						\tiny		%  To minimize the font size
						\begin{minipage}[t]{0.35\columnwidth}		% In order to Wrap text and evoide line exceedings to next cells
							•	Challenging and detached\\
							•	Takes little interest in others\\
							•	Can be seen as insulting or dismissive of others\\
							•	Does not care about other people's feelings or problems\\
							•	Can be manipulative\\
							•	Prefers to be competitive and stubborn\\
							•	Insults and belittles others\\
							•	Manipulates others to get what they want
						\end{minipage}				% In order to Wrap text and evoide line exceedings to next cells
						
					\end{tabular}
					
					& & 
					
					\begin{tabular}{l}
						\tiny		%  To minimize the font size
						\begin{minipage}[t]{0.35\columnwidth}		% In order to Wrap text and evoide line exceedings to next cells
							•	Friendly and compassionate toward others\\ 
							•	Altruist and unselfish\\
							•	Loyal and patient\\
							•	Has a great deal of interest in and wants to help others\\
							•	Feels empathy and concern for other people\\
							•	Prefers to cooperate and be helpful\\
							•	Polite and trustworthy\\
							•	Cheerful and considerate\\
							•	Modest
						\end{minipage}				% In order to Wrap text and evoide line exceedings to next cells

					\end{tabular}
					
					\\
					\hline

					%--------   Neuroticism (N) ---------
					& 		\textbf{Neuroticism (N)} 		\centering		 & 
					\\
					\hline
					
					\begin{tabular}{l}
						\tiny		%  To minimize the font size
						\begin{minipage}[t]{0.35\columnwidth}		% In order to Wrap text and evoide line exceedings to next cells
							•	Emotionally stable\\
							•	Deals well with stress\\
							•	Rarely feels sad or depressed\\
							•	Does not worry much and is very relax\\
							•	Confident and secure\\
							•	Optimist
						\end{minipage}				% In order to Wrap text and evoide line exceedings to next cells

					\end{tabular}
					
					& & 
					
					\begin{tabular}{l}
						\tiny		%  To minimize the font size
						\begin{minipage}[t]{0.35\columnwidth}		% In order to Wrap text and evoide line exceedings to next cells
							•	Anxious of many different things and nervous\\
							•	Experiences a lot of stress\\
							•	Irritable\\
							•	Impulsive and moody\\
							•	Jealous\\
							•	Lack of confidence\\
							•	Self-criticism\\
							•	Oversensitive\\
							•	Instable and insecure\\
							•	Timid\\
							•	Pessimist
						\end{minipage}				% In order to Wrap text and evoide line exceedings to next cells															
					\end{tabular}
					
					\\
					\hline

				\end{tabular}
			\end{adjustbox}
		\end{center}
	}
	
\end{table}

%%%%%%%%%%%%%%%%%%%%%%%%%%%%%%%%%%%%%%%%%%%%%%%%%%%%%%%%%%%%%%%%%%%%%%%%%%%%%%%%%%%%%%%%%%%%%%%%%%%%%%%%%%%%%%%%%
\section{Literature Review}

The history of \textit{psychological researches} in personality goes as far back as Ancient Greece. A number of miscellaneous theories have proposed to explain \textit{what is that makes us who we are}? Some theories are aimed to explain \textit{how personality develops} \cite{ewen2014introduction}, whereas others are concerned with \textit{individual differences in personality} \cite{ashton2013individual}. Reaching maturity, nowadays psychological researches in personality are mainly focused on analyzing the relationship between personality and different human behaviors; such as analyzing the relationship between personality and\footnote{For more attention by computational researchers}:
%----->	Idented Paragraph

\hfill
\begin{minipage}{\dimexpr\textwidth-1cm}
	aggressive and violent behavior \cite{BARLETT2012870}, antisocial behavior \cite{LeCorff2010}, delinquency \cite{yun2017test}, gambling \cite{Reardon2019}, alcohol use \cite{rosenstrom2018prediction}, good citizenship and civic duty \cite{PRUYSERS201999}, emotion regulation strategies \cite{BARANCZUK2019217}, marital instability \cite{mohammadi2018role}, investment and trading performance \cite{CHEN2018}, consumer behaviors \cite{10.1007/978-3-319-47874-6_24}, trustability \cite{Muller2019}, entrepreneurship \cite{LEUTNER201458}, job burnout and job engagement \cite{KIM200996}, employability analysis to find best candidate \cite{10.1007/978-3-319-30927-9_3}, decision making \cite{MENDES201950}, orienting voting choices \cite{VECCHIONE2011737}, political attitudes \cite{JONASON2014181} and preference \cite{ABE201870}, academic/workplace performance \cite{higgins2007prefrontal}, academic motivations and achievements \cite{KOMARRAJU200947}\cite{pozzebon2014major}, learning style \cite{MARCELA20153473} and learning goal orientation \cite{SORIC2017126}, personal goals \cite{REISZ2013699}, forgiveness \cite{walker2017exploring}, subjective well-being \cite{Anglim2016}, and Internet of Things (IoT) \cite{MONTAG2019128}.
	\newline
\end{minipage}

On the other hand, \textit{computational researches} in personality are mainly focused on personality prediction, rather than analyzing its relationship with different behaviors (as mentioned above). Honestly, these researches are still far from the ideal, and they need to be matured through improving the prediction accuracies. However, it is not unexpected that Artificial Intelligence (AI) expedite its maturity quickly, and all of the relevant applications of personality prediction will be tackled masterfully by AI.

Generally contributions in computational personality prediction can be studied from different points of view. The most comprehensive classification could be done based on the source of input information; namely, \textit{text}, \textit{speech}, \textit{video}, \textit{image} and \textit{social media activities}.

%\paragraph{Text:}
\textbf{Text}: written text as a kind of human interaction, would reflect his/her personality. Proving this hypothesis, several studies have focused on personality prediction from the text. Wright and Chin \cite{10.1007/978-3-319-08786-3_21} trained a Support Vector Machine (SVM) to classify the personality of writers in Big Five model, using features such as bag of words, essay length, word sentiment, negation count, and part-of-speech n-grams. They also have investigated the correlation between different features and predictivity of each of the five personality dimensions. In their analysis of written expression, Arjaria et al. \cite{Arjaria2019} questioned a person uniqueness from written text in Big Five model. They proposed a Multi-Label Naïve Bayes (MLNB) classifier to predict the personality of writer. Moreover, they have tested MLNB classifier with a different number of features to find that which written text is formed by which type of trait. Several studies investigating personality prediction from written text, have been carried out on social networks. The study of the structural features of a text in personality prediction was carried out by \cite{8422105}. The authors proposed a bidirectional Long Short Term Memory (LSTM) network, concatenated with a Convolutional Neural Network (CNN) to accomplish the task from YouTube contents. Interestingly, they implemented the evaluations on both short text and long text datasets and proved the suggested models' ability. For the same target, a two level hierarchical deep neural network based on Recurrent CNN structure proposed by Xue et al. \cite{Xue2018}, in order to extract the deep semantic vector representations of each user's text post. Then, they concatenated the vectors with statistical features (like rate of emoticons, rate of capital letters and words, and total number of text posts of each user), to construct the input feature space for traditional regression algorithm to carry out final prediction in Big Five model. Santos et al. \cite{10.1007/978-3-319-64206-2_4} examined that which of the Big Five personality traits are best predicted by different text genres and the needed amount of text for doing the task appropriately. Dandannavar, et al. \cite{8769304} have surveyed personality prediction using social media text. 

\textbf{Speech}: As proved by social psychology, speech includes a lot of information that largely reflect speaker's personality, spontaneously and unconsciously. Mohammadi and Vinciarelli \cite{7344614} showed that it is possible to attribute the Big Five traits to speakers appropriately using prosodic features. They also have compared and analyzed the effect of different prosodic features on prediction. In the same vein, Zhao et al. \cite{10.1007/978-3-319-14445-0_16} performed personality prediction in Chinese. In their investigation into personality prediction, Jothilakshmi et al. \cite{Jothilakshmi2017} proposed a technique based on modeling the relationship between speech signal and personality traits using spectral features. They have used K-Nearest Neighbor and SVM classifier in the Big Five model.

\textbf{Image}: There is no considerable amount of literature on personality prediction from image. Much of the current literature pays particular attention to facial images. The studies in psychology presented thus far provide evidences that, faces play a leading role in daily assessment of human character by others. A recent study by Xu et al. \cite{10.1007/978-3-030-00021-9_54} involved a multi view facial feature extraction model to evaluate the possible correlation between personality traits and face images. Trying to develop a comprehensive personality prediction model based on Support Vector Regression, they proposed 22 facial features. Moreover, they introduced two datasets to investigate the correlation between personality traits and face images. Using the content of Facebook profile pictures, such as facial close ups, facial expressions, and alone or with others, Celli et al. \cite{Celli:2014:API:2647868.2654977} developed a model to extract users' personality. Ferwerda et al. \cite{10.1007/978-3-319-27671-7_71} tried to infer personality traits from the way users manipulate the appearance of their images through applying filters over them, in Instagram. They suggested their method as a new way to facilitate personalized systems.

\textbf{Video}: Undoubtedly personality prediction from pure text, speech or image, despite its successes, ignores some salient information about human characteristic. Recent trends in multimodal data analysis, including visual and audio, have led to a proliferation of studies in personality prediction from videos, which is called \textit{Apparent Personality Analysis} (\textit{APA}). The purpose of APA is to develop methods to find personality traits of users in short video sequences. To develop a personality mining framework, Vo et al. \cite{10.1007/978-3-319-93034-3_51} exploited all the information from videos including visual, audio and textual perspectives, using First Impression dataset and YouTube Personality dataset. Extracting textual, audio and video features, they have used a multimodal mixture density boosting network which combines advanced deep learning techniques to build a multilayer neural network. In order to find those information that apparent personality traits recognition model rely on when making prediction, Gucluturk et al. \cite{Gucluturk_2017_ICCV} performed a number of experiments. They characterize the audio and visual information that drive the predictions. Furthermore, an online application was developed which provides anyone the opportunity to receive feedback on their apparent personality traits. To achieve the APA's purpose, a Deep Bimodal Regression framework was suggested by Zhang et al. \cite{10.1007/978-3-319-49409-8_25}. In the visual modality, they extracted frames from each video and then designed a deep CNN to predict the Big Five traits. They also extracted the log filter bank features from the original audio of each video, in audio modality. Then, a linear regressor was trained to recognize the Big Five traits.

\textbf{Social media activities}: In the light of recent events in personality prediction, there is now some concern about human activities in social media. Besides that, they make it possible to create and share different types of information (text, image, voice, video, links, etc.), exposing their thoughts, feelings and opinions, they allow analyzers to scrutinize the users' different activities, such as likes, visits, mentions, forwards, replies, friends chain, and many others depending on the context. Hima and Shanmugam investigated the correlation between different users' behaviors in social media and the Big Five personality traits \cite{10.1007/978-981-13-9187-3_65}; namely, they found that \textit{extroversion} is correlated with the number of friends or followers, more social media groups, frequent uses of social media, and \textit{neuroticism} is correlated with less use of private messages, sharing more information, spending more time in social media and so on. Chen et al.  \cite{chen2016user} called into question that, is it possible for enterprises to obtain the personality information of customers unconsciously, to employ an effective communication strategy? They showed that, perusing the users' interactions in social media (Facebook), makes it possible to predict the personality of customers. In their analysis, Tadesse et al. examined the presence of structures of social networks and linguistic features to predict users' personality on Facebook \cite{8494744}. Buettner developed a personality-based product recommender framework \cite{Buettner2017}. He analyzed social media data to predict users' personality according to his/her product preferences. A personality prediction system from digital footprint on social media was proposed by Azucar et al.  \cite{AZUCAR2018150}. In addition, they have investigated the impact of different types of digital footprints on prediction accuracy. 

Contributions in computational personality prediction can also be classified differently based on: the output aspect (into \textit{personality predictive systems} and \textit{personality generative systems}), the final target of analysis (into \textit{personality predictor systems} and \textit{personality-based recommender systems}), and other probable classifications. \hyperref[tab: Table2]{Table~\ref*{tab: Table2}}, lists several other contributions in personality prediction for more study. 

The present study will focus on APP methods from `text', without relying on any other probable aspects of input data. Actually considering the fact that the Big Five personality model is on the basis of the lexical hypothesis, the suggested APP methods are basically concentrated on words and terms (specifically, the vocabulary of individuals). Next section, provides a comprehensive description of suggested methods.

\newcolumntype{P}[1]{>{\centering\arraybackslash}p{#1}}	%---> This Row let me to centering the cell content wit P(no "p") same time with sizing the

%\begin{landscape}
\begin{table}[htbp]
	
	\scriptsize

	\begin{center}
		
		\caption{A literature review on APP}
		\label{tab: Table2}
		\begin{adjustbox}{max width=\textwidth}
			\begin{tabular}{@{}p{1cm}@{}p{2.1cm}@{}p{2.1cm}@{}p{3cm}@{}p{2.1cm}@{}p{2cm}@{}p{1.6cm}@{}P{4cm}@{}}
				
				\hline
				\hline
				%-----> First Row
				\tiny
				\cellcolor{gray!15}	\textbf{Ref.}	\centering	&	\cellcolor{gray!15}\tiny	\textbf{Source of the input data}	\centering	&	\cellcolor{gray!15}\tiny	\textbf{Type of input data}	\centering	&	\cellcolor{gray!15}\tiny	\centering	\textbf{System target}	&	\cellcolor{gray!15}\tiny	\textbf{Predictive or Generative}	\centering	&	\cellcolor{gray!15}\tiny	\textbf{Trait Theory}	\centering	&	\cellcolor{gray!15}\tiny	\textbf{Language}	\centering	&	\cellcolor{gray!15}\tiny		\textbf{Technique} 
				
				\\
				\hline
				\hline
				%-----> 2nd Row
				\cite{10.1007/978-3-642-38844-6_29}		\centering	\tiny	&	\tiny Email	\centering		&	\centering	\tiny	Text	&	\centering	\tiny	Personality prediction	&	\centering	\tiny	Predictive	&	\centering	\tiny	Big Five	&	\centering	\tiny	English	  &	
				\begin{minipage}[t]{0.2\columnwidth}	\tiny
					3 learning algorithms, including Joint Model, Sequential Model, and Survival Model
				\end{minipage}
				\\
				\hline
				
				%-----> 3rd Row
				\cite{levitan2016identifying}	\centering		&	\tiny  Dialogue	\centering		&	\tiny	Speech	\centering		&	\tiny	Deception detection	  \centering		&	\tiny	Predictive	\centering		&	\tiny	Big Five	\centering		&	\tiny	English	\centering		&	\tiny Acoustic-prosodic and lexical features as well as 4 machine learning algorithms, including SVM, logistic regression, AdaBoost and random forest

				\\
				\hline
				
				%-----> 4th Row
				\cite{7899604}  \centering   & 	\tiny	Audiovisual data	\centering	&	\tiny	Video	\centering	&	\tiny	Personality prediction	\centering	&	\tiny	Predictive	\centering	&	\tiny	Big Five	\centering	&	\tiny	English		\centering	&	\tiny	Random decision forest

				\\
				\hline
				
				%-----> 5th Row
				\cite{dave2016application}	\centering	&	\tiny	Social media images		\centering	&	\tiny	Image	\centering	&	\tiny	Personality prediction	\centering	&	\tiny	Predictive	\centering	&	\tiny	Big Five	\centering	&	-	\centering	&	\tiny	Convolutional neural networks

				\\
				\hline
				
				%-----> 6th Row
				\cite{8122145}	\centering	&	\tiny	Signature images	\centering	&	\tiny	Image	\centering	&	\tiny	Personality prediction	\centering	&	\tiny	Predictive	\centering	&	\tiny	Big Five	\centering	&	-	\centering	&	\tiny	Back propagation neural network

				\\
				\hline
				
				%-----> 7th Row
				\cite{Souri2018}	\centering	&	\tiny	Social network's profile (Facebook)	\centering	  &			\tiny	Social media activities	\centering	   	&		\tiny	Personality prediction	\centering	&	\tiny	Predictive	\centering	&	\tiny	Big Five	\centering	&	-	\centering	&		\tiny	Boosting decision tree

				\\
				\hline
				
				%-----> 8th Row
				\cite{10.1007/978-3-030-16181-1_46}		\centering	&	\tiny	Social network's profile (Facebook)  \centering	   &	\tiny		Text	\centering	  &	\tiny	Personality prediction	\centering	  &	\tiny	Predictive	\centering	  &	\tiny	5 traits	\centering	  &	\tiny	Spanish	\centering	  &	\tiny	Different algorithms in WEKA

				\\
				\hline
				
				%-----> 9th Row
				\cite{8260763}	\centering	&	\tiny	Social network's profile (Facebook)	  \centering	  	&	\tiny	Social media activities	   \centering	  &	\tiny	Personality prediction	\centering	  &	\tiny	Predictive	\centering	  &	\tiny	Big Five	\centering	  &	-	\centering	  &	\tiny	Least Absolute Shrinkage and Selection Operator algorithm (LASSO)

				\\
				\hline
				
				%-----> 10th Row
				\cite{8104567}		\centering	&	\tiny	Social network's information (Twitter)	\centering	  &	\tiny	Social media activities	 \centering	  &	\tiny	Personality prediction	\centering	  &	\tiny	Predictive	\centering	  &	\tiny	Big Five	\centering	  &	\tiny	Bahasa Indonesia	\centering	  	&	\tiny	SVM and XGBoost algorithm

				\\
				\hline
				
				%-----> 11th Row
				\cite{Gavrilescu2018}	\centering	&	\tiny	Handwriting	 \centering	   &		\tiny	Graphological features	\centering	  &	\tiny	Personality prediction	\centering	  &	\tiny	Predictive	\centering	  &	\tiny	Big Five	\centering	  	&	\tiny	English	\centering	  &	\tiny	Back propagation neural network

				\\
				\hline
				
				%-----> 12th Row
				\cite{10.1007/978-3-319-67401-8_39}		\centering	&	\tiny	Designed character generation tool	\centering	  &	\tiny	Values for 3 personality traits	\centering	  &	\tiny	Generating virtual characters whose physical attributes reflects respected personality traits	\centering	  &	\tiny	Generative	\centering	  &	\tiny	3 personality traits (dominance, agreeableness, trustworthiness)	\centering	  &	-	\centering	  &	\tiny	Linier programming

				\\
				\hline
				
				%-----> 13th Row
				\cite{vernon2014modeling}	\centering	&	\tiny	Not specified	\centering	  &	\tiny	Scores for 3 personality traits		\centering	  &	\tiny	Generating new face illustrations 	\centering	  &	\tiny	Generative	\centering	  &	\tiny	3 personality traits (approachability, dominance, youthful-attractiveness)	\centering	  &	-	\centering	  &	\tiny	A 3 layer neural network in order to synthesize cartoon face-like images

				\\
				\hline
				
				%-----> 14th Row
				\cite{8442031}	\centering	&	\tiny	Questionnaire and Twitter account	\centering	  &	\tiny	Personality traits derived from questionnaires and social media activities	\centering	  &	\tiny	Major recommendation system for students based on their personality	\centering	  &	\tiny	Predictive	\centering	  &	\tiny	MBTI 16 personality types	\centering	  &	\tiny	Arabic	\centering	  &	\tiny	ID3 algorithm

				\\
				\hline
				
				%-----> 14th Row			
				\cite{10.1007/978-3-319-20267-9_25}	\centering	&	\tiny	Users' input	\centering	  &	\tiny	Users personality traits score	\centering	  &	\tiny	Movie recommendation system 	\centering	  &	\tiny	Predictive	\centering	  &	\tiny	Big Five	\centering	  &	-	\centering	  &	\tiny	Gaussian process for personality inference and collaborative filtering for recommendation
				
				\\
				\hline

			\end{tabular}
		\end{adjustbox}
	\end{center}
\end{table}
%\end{landscape}
%%%%%%%%%%%%%%%%%%%%%%%%%%%%%%%%%%%%%%%%%%%%%%%%%%%%%%%%%%%%%%%%%%%%%%%%%%%%%%%%%%%%%%%%%%%%%%%%%%%%%%%%%%%%%%%%%
\section{Methods}
The aim of our study was to enhance the automatic personality prediction from the text, through ensemble modeling. Ensemble modeling causes to the production of better predictive performance compared to each single method, by combining all of the methods. Therefore, we introduced five different basic methods, having various processing level, to be applied in ensemble modeling; including \textit{term frequency vector-based}, \textit{ontology-based}, \textit{enriched ontology-based}, \textit{latent semantic analysis-based}, and \textit{deep learning-based (BiLSTM)} methods.

All the methods in the current study were carried out using \textit{Essays Dataset} \cite{pennebaker1999linguistic}. It contains 2,467 essays and totally 1.9 million words. Essays have been written by psychology students. Then they were asked to reply the Big Five Inventory Questionnaire. Finally, a binary label is attributed to each of the essays in each five personality trait. \hyperref[tab: Table 3]{Table~\ref*{tab: Table 3}} presents the distribution of labels in each of the five traits. Due to the fact that, the general writing style exposes the writer's personality, each of the proposed methods in this study uses Essays Dataset without applying familiar pre-processing activities in natural language processing (like stemming, anaphora resolution, stop words removal, etc.). In other words, in respect to lexical hypothesis and the fact that the personality characteristics of an individual are encoded into his/her language lexicon, we refused to alter the writing style in input essays (as the individual's encoded personality) through common pre-processing activities. Indeed, tokenization was the sole performed pre-processing activity on input essays in all of the suggested APP methods in this study. Furthermore, the Big Five personality model is used in this study.
%-------> Table 3
\begin{table}[htbp]
	\scriptsize
	\begin{center}	
		\caption{The distribution of labels in Essays Dataset}	
		\label{tab: Table 3}
		\begin{tabular}{rcc}\hline\rowcolor{gray!35}
			&	\textbf{True}	&	\textbf{False}			\\\hline\hline
			\textbf{Openness} 	&	1,271	&	1,196	\\\hline
			\textbf{Conscientiousness}	& 	1,253	&	1,214	\\\hline
			\textbf{Extroversion}	&	1,276	&	1,191	\\\hline
			\textbf{Agreeableness}	&	1,310	&	1,157	\\\hline
			\textbf{Neuroticism}		&	1,233	&	1,234	\\\hline
			
		\end{tabular}

	\end{center}

\end{table}

%(((((((((((((((( Base Methods))))))))))))))))
\subsection{Base Methods}
\subsubsection{Term Frequency Vector-Based Method}
\textit{Term frequency} is a weight that is assigned to a term, depending on the number of occurrences of the term in the document. According to the rudimentary principles of information retrieval, term frequency indicates the significance of a particular term within the overall document. The representation of a set of documents as vectors in a common vector space is known as the \textit{vector space model} and is fundamental to a host of information retrieval operations ranging from scoring documents on a query, document classification, and document clustering \cite{manning2010introduction}. Toward this end, we assigned an \textit{essay vector} for each essay in dataset based on the term frequencies, indicating the relative importance of each term in each dimension. In order to find the personality label for an input essay in each of the traits in Big Five model (i.e. O, C, E, A, and N), as Equation \ref{eq:TF-Vector-Based-TRUE} reveals, first we aggregated the cosine similarity between the input essay vector ($e_{i}$) and `True' labeled essays' vectors in Essays Dataset, to achieve the similarity between $e_{i}$ and `True' labeled essays:

\begin{equation}
	\label{eq:TF-Vector-Based-TRUE}
	\centering
	%similarity\_between\_e_{i}\_and\_True\_labeled\_essays  = \sum_{e_{j} \in T} Cosine Similarity (e_{i}, e_{j}) 
	Sim (e_{i} , T)  = \sum_{e_{j} \in T} Cosine Similarity \; (e_{i}, e_{j}) 
\end{equation}
where, $T$ denotes the set of `True' labeled essays' vectors in current personality trait in Essays Dataset. In a same manner, as Equation \ref{eq:TF-Vector-Based-FALSE} shows the similarity between the input essay vector and `False' labeled essays' vectors was calculated:

\begin{equation}
	\label{eq:TF-Vector-Based-FALSE}
	\centering
	%similarity\_between\_e_{i}\_and\_False\_labeled\_essays  = \sum_{e_{j} \in F} Cosine Similarity (e_{i}, e_{j}) 
	Sim (e_{i} , F)  = \sum_{e_{j} \in F} Cosine Similarity \; (e_{i}, e_{j})
\end{equation}
where $F$ stands for the set of `False' labeled essays' vectors in current personality trait in Essays Dataset. At last, as Equation \ref{eq:TF-FinalLabel} reveals the bigger value for similarity, determined the label for the input essay ($Label^{t}(e_{i})$), in current personality trait  $t$, where $t \in \{O, C, E, A, N\}$. That is to say, if the similarity between $e_{i}$ and True labeled essays' vectors is greater than the similarity between $e_{i}$ and False labeled essays' vectors, then the predicted label for $e_{i}$ in current trait will be equal to `True'. \hyperref[Algorithm_1: Term-Frequency Based Method]{Algorithm~\ref*{Algorithm_1: Term-Frequency Based Method}} shows the pseudocode of term frequency vector-based APP.

\begin{equation}
	\label{eq:TF-FinalLabel}
	Label^{t}(e_{i}) =
	\begin{cases}
		True & \text{, $if \; \; Sim (e_{i} , T) \geq Sim (e_{i} , F)$}\\
		False & \text{, \textit{otherwise}}\\
	\end{cases}
\end{equation}

\begin{algorithm}
	\caption{Term frequency vector-based APP algorithm}
	\label{Algorithm_1: Term-Frequency Based Method}
	\scriptsize
	
	\begin{algorithmic}[1]
		\State Create essay vector $e$ for each essay in Essays Dataset
		\MyFor{trait $t \in \{O, C, E, A, N\}$}
		\MyFor{$e_{i} \in $ Essays Dataset}
		\MyFor{$e_{j} \in $ Essays Dataset where $i \neq j$}
		\IF{the gold label in trait $t$ for $e_{j}$ == `\textit{True}'}
		\STATE similarity\_between\_$e_{i}$\_and\_True\_labeled\_essays + = Cosine Similarity ($e_{i}$, $e_{j}$)
		\ELSIF{the gold label in trait $t$ for $e_{j}$ == `\textit{False}'}
		\STATE similarity\_between\_$e_{i}$\_and\_False\_labeled\_essays + = Cosine Similarity ($e_{i}$, $e_{j}$)
		\ENDIF
		\EndMyFor
		\IF{similarity\_between\_$e_{i}$\_and\_True\_labeled\_essays $>$  similarity\_between\_$e_{i}$\_and\_False\_labeled\_essays}
		\STATE the predicted label for $e_{i}$ in trait $t$ = `\textit{True}'
		\ELSE
		\STATE the predicted label for $e_{i}$ in trait $t$ = `\textit{False}'
		\ENDIF
		\EndMyFor
		\EndMyFor
	\end{algorithmic}
\end{algorithm}

%-------------------------------------------------------------------------------------------------------------------------------------------------------------------
\subsubsection{Ontology-Based Method}
\label{sec:Ontology-Based-Method}
In the context of computer and information science, \textit{ontology} is a set of representational primitives, intended for modeling knowledge about individuals, their attributes, and their relationships to other individuals \cite{Gruber2009}. As a way of knowledge representation for machine, ontology groups all of the individuals into sets of cognitive synonyms called \textit{synsets}, each expressing a distinct concept. Synsets are interlinked through conceptual-semantic and lexical relations. In an effort to automatic personality prediction, we hypothesize that the used synsets' networks by writers are able to expose his/her personality traits in texts. To reach that, first we obtained two sets for each personality trait ($t$) in the Big Five model, including:
\begin{enumerate}[label=(\roman*)]
	\item $T_{indv}(t)$ which refers to the set of the extracted individuals and their related synsets from `True' labeled essays in trait $t$ in Essays Dataset, where $t \in \{O, C, E, A, N\}$;
	\item $F_{indv}(t)$ which refers to the set of the extracted individuals and their related synsets from `False' labeled essays in trait $t$ in Essays Dataset.
\end{enumerate}

Each of the extracted individuals and synsets has an \textit{importance} value that is equal to one at its first occurrence. During extraction, the reoccurrence of each individual or sysnet in next times, increases its importance (one for each). In other words, as Equations \ref{eq:Ontology-Vector-Based-TRUE} and \ref{eq:Ontology-Vector-Based-FALSE} reveal, each individual or synset has an importance value equal to its number of occurrences throughout the `True' or `False' labeled essays (namely, $freq (x)$).

\begin{equation}
	\label{eq:Ontology-Vector-Based-TRUE}
	\centering
	importance (x) = freq (x) \; \; \; \;, \; \; \; \forall x \in  T_{indv}(t) 
\end{equation}

\begin{equation}
	\label{eq:Ontology-Vector-Based-FALSE}
	\centering
	importance (x) = freq (x) \; \; \; \;, \; \; \;  \forall x \in  F_{indv}(t)
\end{equation}

Then, for a given input essay ($e_{i}$), we extracted all of the individuals and their related synsets in a same manner. Next, for predicting the input essays' labels in each of the five traits, as Equations \ref{eq:Ontology-True-Final-Similarity} and \ref{eq:Ontology-False-Final-Similarity} denote, we aggregated the importance of common individuals and synsets between the input essay and `True' labeled essays, as well as `False' labeled essays:

\begin{equation}
	\label{eq:Ontology-True-Final-Similarity}
	Sim (e_{i} , T_{indv}) = \sum_{x \in T_{intsec}} importance (x)	\; \; \; \;, \; \; \; \forall e_{i} \in Essays Dataset
\end{equation}

\begin{equation}
	\label{eq:Ontology-False-Final-Similarity}
	Sim (e_{i} , F_{indv}) = \sum_{x \in F_{intsec}} importance (x)	\; \; \; \;, \; \; \; \forall e_{i} \in Essays Dataset
\end{equation}
\underline{where,}
\begin{equation}
	T_{intsec} = (individuals \; and \; synsets \; in \; e_{i}) \cap T_{indv}
\end{equation}

\begin{equation}
	F_{intsec} = (individuals \; and \; synsets \; in \; e_{i}) \cap F_{indv}
\end{equation}

Finally, as the Equation \ref{eq:Ontology-FinalLabel} denotes, the bigger similarity value, determined the label for input essay in the current trait. \hyperref[Algorithm_2: Ontology-based method APP]{Algorithm~\ref*{Algorithm_2: Ontology-based method APP}}, represents the pseudocode of ontology-based APP. In order to do so, we used WordNet 2.0 \cite{fellbaum2005wordnet} ontology. It contains 152,059 unique individuals and 115,424 synsets. 

\begin{equation}
	\label{eq:Ontology-FinalLabel}
	Label^{t}(e_{i}) =
	\begin{cases}
		True & \text{, $if \; \; Sim (e_{i} , T_{indv}) \geq Sim (e_{i} , F_{indv})$}\\
		False & \text{, \textit{otherwise}}\\
	\end{cases}
\end{equation}

\begin{algorithm}
	\caption{Ontology-based APP algorithm \newline \footnotesize ignore * and ** here, they will be added in next (enriched ontology-based) method}
	\label{Algorithm_2: Ontology-based method APP}
	\scriptsize
	\begin{algorithmic}[1]
		
		\MyFor{trait $t \in \{O, C, E, A, N\}$}
		\State \parbox[t]{420pt}{extract all of the existing individuals and their related synsets from `\textit{True}' labeled essays in trait $t$ from Essays Dataset, and call it \textit{True\_labeled\_essays\_individuals\_in\_$t$}; then set an \textit{importance} value for each of them considering the `\textit{importance assignment rule}'
		\newline}
		\State *
		\State \parbox[t]{420pt}{extract all of the existing individuals and their related synsets from `\textit{False}' labeled essays in trait $t$ from Essays Dataset, and call it \textit{False\_labeled\_essays\_individuals\_in\_$t$}; then set an \textit{importance} value for each of them considering the `\textit{importance assignment rule}'
		\newline}
		\State ** 		\newline
		\STATEx \hskip4.0em \parbox[t]{410pt}{\textbf{\textit{importance assignment rule}:} set an importance value for each of the individuals and synsets equal to 1 on their first occurrences, and add 1 to them for each of their next reoccurrences
		\newline
		\newline}
		\MyFor{essay $e_{i} \in $ Essays Dataset}
		\State \parbox[t]{410pt}{extract all of the existing individuals and their related synsets from $e_{i}$
		\newline}
		\State \parbox[t]{420pt}{set \textit{similarity\_between\_$e_{i}$\_and\_True\_labeled\_essays} equal to the aggregation of common individuals' and synsets' \textit{importance} values between $e_{i}$ and \textit{True\_labeled\_essays\_individuals\_in\_$t$} (extracted in step 2)
		\newline}
		\State \parbox[t]{420pt}{set \textit{similarity\_between\_$e_{i}$\_and\_False\_labeled\_essays} equal to the aggregation of common individuals' and synsets' \textit{importance} values between $e_{i}$ and \textit{False\_labeled\_essays\_individuals\_in\_$t$} (extracted in step 4)
		\newline}
		\IF{\textit{similarity\_between\_$e_{i}$\_and\_True\_labeled\_essays} $>$  \textit{similarity\_between\_$e_{i}$\_and\_False\_labeled\_essays}}
		\STATE the predicted label for $e_{i}$ in trait $t$ = `\textit{True}'
		\ELSE
		\STATE the predicted label for $e_{i}$ in trait $t$ = `\textit{False}'
		\ENDIF

		\EndMyFor
		\EndMyFor
		
	\end{algorithmic}
\end{algorithm}

%-------------------------------------------------------------------------------------------------------------------------------------------------------------------
\subsubsection{Enriched Ontology-Based Method}
There is an undeniable correlation between personality and individual differences in word use. Yarkoni investigated and presented the most correlated words (both positive and negative) with five personality traits in the Big Five model \cite{YARKONI2010363}. For instance, he showed that \textit{`restaurant'} (0.21) and \textit{`drinks'} (0.21) are positively, and \textit{`cat'} (-0.2) and \textit{`computer'} (-0.19) are negatively correlated with extroversion. In the previous method, we extracted all of the individuals and their synsets from `True' and `False' labeled essays in each personality trait, separately. Here, intended to enrich the previous method, we added the most correlated words (as listed in \cite{YARKONI2010363}) and their synsets up to the previously extracted sets correspondingly in each personality trait. Then similar to the ontology-based method in order to predict an input essay’s labels in each trait, the importance values of existed individuals and synsets were aggregated; both in `True' labeled essays and `False' labeled essays. At last, the bigger similarity value determined the label for the current personality trait. All of the stated equations in section \ref{sec:Ontology-Based-Method}, are also valid here. The sole difference is that, the sets of $T_{indv}(t)$ and $F_{indv}(t)$, were enriched.

Placing the steps bellow in \hyperref[Algorithm_2: Ontology-based method APP]{Algorithm~\ref*{Algorithm_2: Ontology-based method APP}}, will extend the ontology-based APP algorithm to enriched ontology-based APP.

\begin{algorithm}
	
	\scriptsize
	\begin{algorithmic}
		\State 3: *  add \textit{positively correlated} words in $t_{i}$ to the \textit{True\_labeled\_essays\_individuals\_in\_$t_{i}$} and set their \textit{importance} values considering the \textit{importance assignment rule} 
		\STATEx
		\State 5: **  add \textit{negatively correlated} words in $t_{i}$ to the \textit{False\_labeled\_essays\_individuals\_in\_$t_{i}$} and set their \textit{importance} values considering the \textit{importance assignment rule}
	\end{algorithmic}
\end{algorithm}

%-------------------------------------------------------------------------------------------------------------------------------------------------------------------
\subsubsection{Latent Semantic Analysis (LSA)-Based Method}
\textit{Latent Semantic Analysis (LSA)} is a method for extraction and representation of the contextual meaning of words by statistical computations applied to a large corpus of text \cite{10108001638539809545028}. The main idea behind LSA is that, the presence or non-presence of the words in different contexts are considered as the basis of mutual limitations, that significantly determines the similarity of meaning of words and phrases \cite{landauer2013handbook}. LSA is a method in vector space that is considerably able to identify and extract the semantic relations those are formed in the mind of writers during writing. The capability of LSA in terms of representing human knowledge, has been approved in a variety of ways \cite{landauer2013handbook}. LSA is a mathematical-statistical method composed of two basic steps: the first step involves the representation of input text in the form of a matrix whose rows are allocated to each of the unique words in the text, and columns are allocated to sentences, passage or any other text units or contexts (such as paragraph and document or essay). Choosing essay as the considered text unit, we will have a \textit{term$\times$essay (t$\times$e)} matrix (hereafter \textit{A}), whose cells are equal to the frequency of the terms in the corresponding essay. The second step of LSA is applying the mathematical operator calling \textit{Singular Value Decomposition (SVD)} over the matrix obtained from the previous step. Applying SVD to \textit{A}, decomposes it into three matrices; namely, \textit{U}, \textit{$\Sigma$} and \textit{V}.

\begin{equation}
	\label{eq: euation1}
	A_{t\times e} = U_{t\times c} \;  \Sigma_{c\times c} \; V^T_{c\times e} 
\end{equation}

Where \textit{U} stands for a \textit{term$\times$concept (t$\times$c)} and column-orthogonal matrix whose columns are called left singular values; \textit{$\Sigma$=diag{($\sigma_1$, $\sigma_2$, …. , $\sigma_c$)}}  is a diagonal and \textit{concept$\times$concept (c$\times$c)} matrix whose main diagonal elements are the \textit{eigenvalues} of \textit{A}. These values have been sorted in descending order in main diagonal of the matrix; $V^T$ denotes an orthogonal \textit{concept$\times$essay (c$\times$e)} matrix whose columns are called right singular values.

Matrices obtained from applying SVD to \textit{A} can be interpreted as follows; $U_{t \times c}$ is a matrix in which each column identifies a skilfully extracted concept (or topic) from the text by SVD. The rows in \textit{U}, same as the \textit{A}, are allocated to the unique terms. The cells' values in \textit{U} emphasize the weight (importance) of each term in the corresponding concept. Besides, \textit{$\Sigma_{c \times c}$} is a diagonal matrix with descending values in the main diagonal, each of which values emphasize the importance of concepts in the essay. Not all the cells in the main diagonal in \textit{U} have nonzero values; just a number of them which is called \textit{rank(A)}, have nonzero values. As a consequence, applying SVD to \textit{A} practically causes to a dimensionality reduction and trivial concepts removal. It is approved that such dimensionality reduction leads to better approximations for human cognitive behavior. Finally, \textit{$V^T_{c \times e}$} indicates a new representation for essays based on extracted concepts by SVD, rather than the original terms or words.

For the purpose of APP, first for each trait $t$ we composed the \textit{A} matrix for `True' labeled essays, namely $A^{True Labeled}_{t \times e}$, as well as the \textit{A} matrix for `False' labeled essays, namely $A^{False Labeled}_{t \times e}$. Therefor, ten matrices were obtained (two for each personality trait), in which all of them have a dimensionality equal to \textit{t $\times$e}. 
Then, we applied SVD and decomposed all of the ten matrices to \textit{U}, $\Sigma$ and \textit{V} corresponding matrices, as shown in Equations \ref{eq:LSA-SVD-True} and \ref{eq:LSA-SVD-False}.

\begin{equation}
	\label{eq:LSA-SVD-True}
	A^{True Labeled}_{t \times e} = U^{True Labeled}_{t\times c} \;  \Sigma^{True Labeled}_{c\times c} \; V^{True Labeled}_{c\times e} 
\end{equation}

\begin{equation}
	\label{eq:LSA-SVD-False}
	A^{False Labeled}_{t \times e} = U^{False Labeled}_{t\times c}  \; \Sigma^{False Labeled}_{c\times c} \; V^{False Labeled}_{c\times e}
\end{equation}

Next, in each personality trait, the \textit{U} and $\Sigma$ decomposed matrices were selected for both `True' labeled, and `False' labeled essays. As revealed in Equations \ref{eq:LSA-BMatrix-True} and \ref{eq:LSA-BMatrix-False}, multiplying $U_{t \times c}$ and $\Sigma_{c \times c}$, a $B_{t \times c}$ was achieved for both `True' labeled, and `False' labeled essays, which emphasizes the importance of each term in each concept. Furthermore, multiplying \textit{U} and $\Sigma$ will causes to remove less importance terms (by attributing zero values).

\begin{equation}
	\label{eq:LSA-BMatrix-True}
	B^{True Labeled}_{t \times c} = U^{True Labeled}_{t\times c}  \; \Sigma^{True Labeled}_{c\times c}
\end{equation}

\begin{equation}
	\label{eq:LSA-BMatrix-False}
	B^{False Labeled}_{t \times c} = U^{False Labeled}_{t\times c} \;  \Sigma^{False Labeled}_{c\times c}
\end{equation}

Once giving an input essay ($e_{i}$), a \textit{t $\times$1} vector was produced. After that, in order to acheive the similarity between $e_{i}$ and `True' labeled essays as well as `False' labeled essays, as shown in Equation \ref{eq:LSA-Similarity-True} and \ref{eq:LSA-Similarity-False}, we aggregated the column wise cosine similarity between the input essay vector and \textit{B} matrices, both for `True' labeled and `False' labeled essays: 

\begin{equation}
	\label{eq:LSA-Similarity-True}
	Sim (e_{i} , T) = \sum_{e_{j} \in B^{True Labeled}} Cosine Similarity (e_{i}, e_{j})
\end{equation}

\begin{equation}
	\label{eq:LSA-Similarity-False}
	Sim (e_{i} , F) = \sum_{e_{j} \in B^{False Labeled}} Cosine Similarity(e_{i}, e_{j})
\end{equation}
where, $T$ denotes the `True' labeled and $F$ denotes the `False' labeled essays in current personality trait.

Finally, as Equation \ref{eq:LSA-FinalLabel} reveals the bigger value for similarity, determined the label for the input essay in trait $t$. \hyperref[Algorithm_3: LSA_Based APP]{Algorithm~\ref*{Algorithm_3: LSA_Based APP}}, details the pseudocode of the LSA-based APP.

\begin{equation}
	\label{eq:LSA-FinalLabel}
	Label^{t}(e_{i}) =
	\begin{cases}
		True & \text{, $if \; Sim (e_{i} , T) \geq Sim (e_{i} , F)$}\\
		False & \text{, \textit{otherwise}}\\
	\end{cases}
\end{equation}

\begin{algorithm}
	\caption{LSA-based APP algorithm}
	\label{Algorithm_3: LSA_Based APP}
	\scriptsize
	\begin{algorithmic}[1]
		
		\MyFor{trait $t \in \{O, C, E, A, N\}$}
		\State \parbox[t]{420pt}{create $term \times essay$ matrix for all of the `\textit{True}' labeled essays in trait $t$ in Essays Dataset and call it $A^{True Labeled}_{term\times essay}$ in which whose: rows are unique terms in Essays Dateset, columns are `\textit{True}' labeled essays' vectors in $t$, and cells are the frequency of terms in corresponding essay
		\newline}
		\State \parbox[t]{420pt}{create $term \times essay$ matrix for all of the `\textit{False}' labeled essays in trait $t$ in Essays Dataset and call it $A^{False Labeled}_{term\times essay}$ in which whose: rows are unique terms in Essays Dateset, columns are `\textit{False}' labeled essays' vectors in $t$, and cells are the frequency of terms in corresponding essay
		\newline}
		\State \parbox[t]{420pt}{apply SVD operator to the matrices $A^{True Labeled}_{term\times essay}$ and $A^{False Labeled}_{term\times essay}$ and decompose them
		\newline
		\newline
		$A^{True Labeled}_{term\times essay} = U^{True Labeled}_{term\times concept} \; \; \Sigma^{True Labeled}_{concept\times concept} \; \; V^{True Labeled}_{concept\times essay} $
		\newline
		\newline
		$A^{False Labeled}_{term\times essay} = U^{False Labeled}_{term\times concept} \; \; \Sigma^{False Labeled}_{concept\times concept} \; \; V^{False Labeled}_{concept\times essay} $
		\newline
		\newline
		}
		
		\State \parbox[t]{420pt}{create $B_{term \times concept}$ matrices using decomposed matrices
		\newline
		\newline
		$B^{True Labeled}_{term\times concept} = U^{True Labeled}_{term\times concept} \; \;  \Sigma^{True Labeled}_{concept\times concept}$
		\newline
		\newline
		$B^{False Labeled}_{term\times concept} = U^{False Labeled}_{term\times concept} \; \;  \Sigma^{False Labeled}_{concept\times concept}$
		\newline
		\newline
		}
	
		\State \parbox[t]{420pt}{create $term \times 1$ essay vectors $e_{i}$ for all of the essays in Essays Dataset
		\newline}
		\MyFor{essay $e_{i} \in $ Essays Dataset}
		\State \parbox[t]{410pt}{\textit{similarity\_between\_$e_{i}$\_and\_True\_Labeled\_essays} = aggregation of column wise cosine similarities between $e_{i}$ and $B^{True Labeled}$
		\newline}
		\State \parbox[t]{410pt}{\textit{similarity\_between\_$e_{i}$\_and\_False\_Labeled\_essays} = aggregation of column wise cosine similarities between $e_{i}$ and $B^{False Labeled}$
		\newline}
		
		\IF{\textit{similarity\_between\_$e_{i}$\_and\_True\_Labeled\_essays} $>$  \textit{similarity\_between\_$e_{i}$\_and\_False\_Labeled\_essays}}
		\STATE the predicted label for $e_{i}$ in trait $t$ = `\textit{True}'
		\ELSE
		\STATE the predicted label for $e_{i}$ in trait $t$ = `\textit{False}'
		\ENDIF

		\EndMyFor
		\EndMyFor
		
	\end{algorithmic}
\end{algorithm}

\subsubsection{Deep Learning-Based Method}
As the fifth method for APP, we suggested a deep learning \textit{Bidirectional Long Short Term Memory (BiLSTM)} network. Unlike traditional neural networks that cope with inputs independently, \textit{Recurrent Neural Networks (RNN}) take into account a set of previous inputs. This property competently makes RNNs capable of learning from sequential data, like text documents. LSTM as a special kind of RNNs, benefits understanding of previous words (past information) to understand each word (in forward direction), Whilst BiLSTM benefits both understandings of previous and next words (past and future information) in forward and backward directions respectively. It looks much like humans do.

\begin{figure}
	\centering
	\includegraphics[height=6cm]{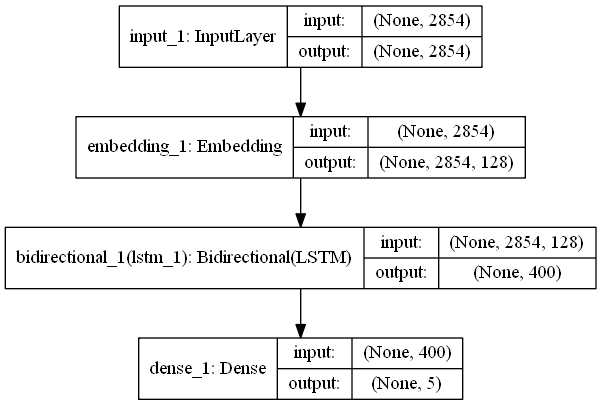}
	\caption{Summary of the deep learning model}
	\label{fig: Figure 1}
\end{figure}

First of all, some preprocessing steps like tokenization, data cleaning and lowercasing were performed on essays’ texts. Next, each unique tokenized word was mapped to an integer id and accordingly, the list of words in each essay text was converted to a list of ids and were post padded to the maximum essay length. In consequence, 2,467 essay vectors with the same lengths (2,854) were produced. The proposed network architecture includes four layers: an input layer, an embedding layer with 128 neurons, a BiLSTM layer with 200 neurons, and a dense layer with five neurons (one for each personality trait). \hyperref[fig: Figure 1]{Figure~\ref*{fig: Figure 1}} shows a summary of the model. It must be noted that, we used early stopping to avoid overfitting with tenfold cross-validation in our experiments. Besides, the designed network performs multi-label text classification for five personality traits, in parallel.

\subsection{Ensemble Modeling}
The main idea behind \textit{ensemble modeling} is to improve the overall predictive performance through combining the decisions from multiple models \cite{vanRijn2018}. Actually, it takes into account the multiple diverse base models' votes to predict an outcome, rather than a single model (much like using the wisdom of the crowd in making a prediction). As a consequence of manipulating independent and diverse base models, the generalization error of prediction will be decreased \cite{kotu2018data}, that is a great incentive to use ensemble modeling. Among different ensemble methods, \textit{Bagging (Bootstrap Aggregation)} \cite{breiman1996bagging}, \textit{Boosting} \cite{FREUND1997119}, and \textit{Stacking} \cite{WOLPERT1992241} are more common. The first two ones are often used to combine homogeneous base methods \cite{10.1145/3093241.3093262} aimed to decrease variance and reduce the bias respectively, whilst the third one is often used to combine heterogeneous base methods \cite{KANG201535} aimed to improve predictions. Experiments show that, applying heterogeneous base methods will improve the ensemble modeling efficiency \cite{van2018online}. Regarding our different five base methods, a stacking-based ensemble modeling was suggested to accomplish APP objectives.

\hyperref[Algorithm_4: stacking]{Algorithm~\ref*{Algorithm_4: stacking}}, presents the stacking algorithm which was used for the purpose of ensemble modeling here. As can be seen, the algorithm contains three sequential steps: it commences by making decision about base classification methods as well as a meta-model, in the first step. Aforementioned five suggested APP methods, were selected as the five base classification methods, and a Hierarchical Attention Network (HAN) \cite{yang2016hierarchical} was selected as the meta-model, also. During the second step, base methods are responsible for making predictions, that are not the final predictions but they have the essential role in making them. Actually, base methods’ predictions (as the wisdom of the crowd) are then used to train the meta-model in the third step. At last, the meta-model will make the final predictions on input essays.

\begin{algorithm}
	\caption{Stacking algorithm}
	\label{Algorithm_4: stacking}
	\scriptsize
	\begin{algorithmic}[1]
		
		\STATEx \textbf{Step 1: initial decisions}
		\
		\State \hskip2.0em \parbox[t]{400pt}{
			\textit{specification of M base classification methods; five base methods were selected including term frequency vector-based, ontology-based, enriched ontology-based, latent semantic analysis (LSA)-based, and deep learning-based (BiLSTM) methods}\strut}
		\State \hskip2.0em \textit{specification of a meta-model; a Hierarchical Attention Network (HAN) was selected}

		\STATEx \textbf{Step 2: making prediction using base classification methods}
		%\Indent
		\State \hskip2.0em \textit{run M base classification methods}
		\State \hskip2.0em \parbox[t]{400pt}{
			\textit{using k-fold cross-validation, collect prediction results from each of the M base classification methods in each of the five personality traits for all of the essays in Essays Dataset}\strut}
		
		\State \hskip2.0em \parbox[t]{400pt}{
			\textit{create an N$\times$M matrix for each five personality trait using predicted values from each of the M base methods; thereby five matrices will be achieved, one for each personality trait (N denotes the number of essays in Essays Dataset)}\strut}
		%\EndIndent
		
		\STATEx \textbf{Step 3: setting up the ensemble model}
		
		\State \hskip2.0em \parbox[t]{400pt}{
			\textit{train the meta-model (HAN) using the obtained matrices and essays’ actual labels (creating the ensemble model)}\strut}
		\State \hskip2.0em \textit{generate the ensemble predictions}

	\end{algorithmic}
\end{algorithm}

The major objectives of the Hierarchical Attention Network or HAN, are to obtain two fundamental insights about document structure \cite{yang2016hierarchical}: firstly, achieving a hierarchical representation of documents (hierarchy); and secondly, achieving context-dependent meaning or importance of different words and sentences (attention). The idea behind the former objective is that, since documents utilize a hierarchical structure, namely words form sentences and sentences form documents, a hierarchical representation of documents will also be helpful. The idea that different words and sentences in a document, context dependently convey different information and importance, forms the main idea behind the next objective. Paying attention to those words and sentences which more contribute to classification decisions, would be so valuable and may result in better performance \cite{10.1145/2661829.2661935}. They can be regarded as convincing incentives to use HAN as a meta-model in ensemble modeling of APP.

\begin{figure}
	
	\centering
	
	\includegraphics[width=\textwidth]{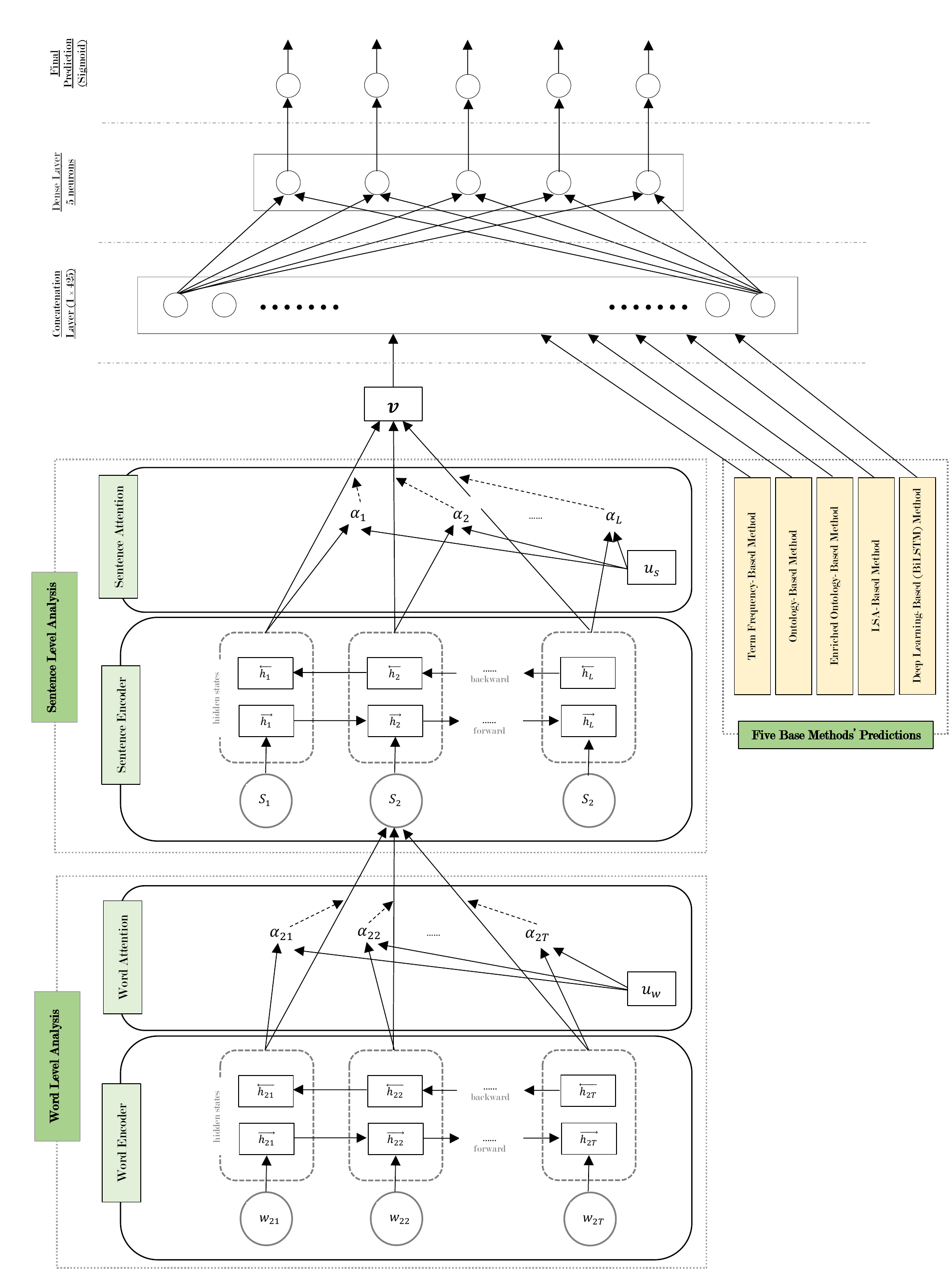}
	
	\caption{the architecture of a Hierarchical Attention Network (HAN) combined with base methods’ predictions to carry out stacking in document level classification.\newline
		Where \textit{L} is equal to the number of sentences (\textit{s}) in essay; each sentence $s_i$ contains $T_i$ words; $w_{it}$ where $t \in [1, T]$ denotes the words in the $i^{th}$ sentence; $h$ implies the hidden states that serves as memory cells to track the state of sequences in forward ($\protect\overrightarrow{h}$) and backward ($\protect\overleftarrow{h}$) directions, in  order to incorporate contextual information; $\alpha$ stands for word/sentences attentions which uses $u_w$/$u_s$ as the word/sentence level context vector; \textit{v} denotes the document level vector of each essay.
	}
	\label{fig: Figure 2}
\end{figure}

\hyperref[fig: Figure 2]{Figure~\ref*{fig: Figure 2}}, depicts the general architecture of a HAN which is combined with base methods’ predictions to carry out the stacking algorithm. The HAN produces a document level vector (\textit{v}) for each of the essays in Essays Dataset, step by step from the word level vectors. An overview of its stepwise operation includes: \textit{i)} \textit{\textbf{word encoder}} that embeds the words to vectors, incorporating the contextual information of them (in forward and backward directions); \textit{ii)} \textbf{\textit{word attention}} that extracts (pays attention) those words that are important to the meaning of the sentence, due to the fact that they have unequal contributions in sentence meaning; \textit{iii)} \textit{\textbf{sentence encoder}} that produces document vector using sentence vectors, incorporating contextual information of sentences in the same manner; \textit{iv)} \textit{\textbf{sentence attention}} that rewards those sentences which play determining role in document classification. The final output of HAN is a document level vector (\textit{v}) for each essay that conveys all the information of sentences in the essay (for more details about HAN, please refer to \cite{yang2016hierarchical}). Practically, \textit{v} is considered as the meaningful extracted features that are resulted from the meta-model. 

With the purpose of ensemble modeling, the HAN were trained using Essays Dataset. 80\% of the essays were used to train the model, and 20\% were used to test it. Following the production of document level vectors for essays, they were fed to a dense layer in combination with suggested five base methods predictions, in order to train the meta-model to produce the final predictions. Finally, the sigmoid activation function determines the final prediction of essays classification in each of the five personality traits. 

The fusion of the outputs from five various base methods into the dense vector of \textit{v} is applied by concatenation of them. To be specific, the \textit{v} vector (the output of HAN) that is in the shape of 1$\times$400, and the outputs of five base methods, each of which is a 1$\times$5 binary vector (zeros and ones; zero for \textit{False} label and one for \textit{True} label in corresponding personality trait), are concatenated into a single vector. After the concatenation, a 425 dimensional vector will be produced. Applying a dense layer with 5 neurons on the concatenated vector, that each of them come to an end with a sigmoid activation function (in order to ensure the probability independence of the five distinct personality traits) at the output layer, yields a multi-label output. Furthermore, the weights of this layer are automatically adjusted by back-propagation using Adam optimizer (with the learning rate of 1e-5). The weights will rule a coefficient model on the 425 dimensional concatenated vector having a look at the essay’s text (the first 400 elements), as well as an aid from the five base methods’ predictions (the last 25 elements), to produce the final output.

%%%%%%%%%%%%%%%%%%%%%%%%%%%%%%%%%%%%%%%%%%%%%%%%%%%%%%%%%%%%%%%%%%%%%%%%%%%%%%%%%%%%%%%%%%%%%%%%%%%%%%%%%%%%%%%%%
\section{Results}
\subsection{Evaluation Measures}
\textit{Precision}, \textit{recall}, \textit{f-measure}, and \textit{accuracy} are the most well-known evaluation measures in binary classification systems (\cite{manning2010introduction} for more study). Generally, there are two sets of labels for documents that have determining role in evaluation measures, containing \textit{gold standard} and \textit{ground truth}. The former refers to the set of reference labels for the documents, as the most accurate benchmark, and the latter refers to the set of system assigned (predicted) labels for documents. Moreover, there are two another determining notions, namely \textit{relevant} and \textit{non-relevant} documents. Considering the user's needed information, relevant documents are those that meet the needs and vice versa. 

Practically, there are four possible combinations of actual labels and system assigned labels (observations), as shown in \hyperref[tab: Table 4]{Table~\ref*{tab: Table 4}}: \textit{true positive} or \textit{TP} (the number of retrieved items that are relevant), \textit{false positive} or \textit{FP} (the number of retrieved items that are non-relevant), \textit{false negative} or \textit{FN} (the number of not retrieved items that are relevant), and \textit{true negative} or \textit{TN} (the number of not retrieved items that are non-relevant). Respecting the contents of \hyperref[tab: Table 4]{Table~\ref*{tab: Table 4}}, a better understanding about measures and the idea behind each of them, could be achieved. 

%-------> Table 4
\newcolumntype{P}[1]{>{\centering\arraybackslash}p{#1}}	%---> This Row let me to centering the cell content wit P(no "p") same time with sizing the cell
\begin{table}[htbp]
	\tiny
	\begin{center}	
		\caption{Possible combinations of reference labels and system assigned labels}	
		\label{tab: Table 4}
		\begin{tabular}{P{2.7cm}P{2.3cm}P{0.6cm}|P{2.0cm}|P{2.3cm}|}\cline{4-5}
			\hhline{~|~|~|~|-|}
			
			\multirow{3}{*}{} &	\multirow{3}{*}{} &	\multirow{3}{*}{} &	\multicolumn{2}{c|}{\cellcolor{orange!45}	\textbf{Gold Standard} (Reference Labels)}	\\
			
			\cline{4-5}
			\hhline{~|~|~|-|-|}
			&	&	&	\cellcolor{black!15}	Relevant Items	& 	\cellcolor{black!15}	Non-Relevant Items\\
			
			\cline{4-5} \cline{4-5}
			\hhline{~|~|~|-|-|}
			&	&	&	 \cellcolor{yellow!20}	\textbf{True}	& 	\cellcolor{yellow!15}	\textbf{False}\\
			\hhline{~|~|~|-|-|}
			\cline{1-5}
			\hhline{~|~|-|~|~|}
			%\hhline{~|~|~|=|=|}
			%\hhline{---==}
			\multirow{2}{*}{\makecell{\textbf{Ground Truth}\cellcolor{orange!45} \\ \cellcolor{orange!45}(System Assigned Labels)}} &		\cellcolor{black!15}	Retrieved Items	&	 \cellcolor{yellow!20}	\textbf{True}	&	\textit{TP}	&	\textit{FP}\\
			
			\cline{2-5}
			\hhline{~|-|-|~|~|}
			&	\cellcolor{black!15}	Not Retrieved Items 	&	\cellcolor{yellow!20}	\textbf{False}	&	\textit{FN}	&	\textit{TN}\\\hline
			
		\end{tabular}

	\end{center}

\end{table}

\textit{Precision} is equal to the proportion of the number of retrieved documents that are relevant. As Equation \ref{eq: Equation 2} reveals, high values for precision, denote to the low false positive rates. Actually, precision is intended to reply the question that \textit{``what proportion of all the items that are labeled by the system, are correctly labeled''}? This is why it is called `precision'.
\begin{equation}
\scriptsize
\label{eq: Equation 2}
\begin{aligned}
Precision =&  \frac{\# \text{\textit{relevant items reteived}}}{\# \text{\textit{retreived items}}}
\\
=&  \frac{\# \text{\textit{system predicted True label items}}}{\# \text{\textit{total system predicted True items (Ground Truth)}}}
\\
=&  \frac{TP}{TP+FP}
\end{aligned}
\end{equation}
\textit{Recall} is equal to the proportion of relevant documents that are retrieved. Considering Equation \ref{eq: Equation 3}, recall mainly tries to reply the question that \textit{``what proportion of expected items, are correctly labeled by the system''}? This is why it is called `recall'.
\begin{equation}
\scriptsize
\label{eq: Equation 3}
\begin{aligned}
Recall =&  \frac{\# \text{\textit{relevant items reteived}}}{\# \text{\textit{relevant items}}}
\\
= & \frac{\# \text{\textit{system predicted True label items}}}{\# \text{\textit{all items in Gold Standard}}}
\\
=&  \frac{TP}{TP+FN}
\end{aligned}
\end{equation}
It should be noted that, precision and recall are not individually appropriate for evaluating the system performance. Specifically, there would be systems with high precision but low recall, and vice versa. Addressing this matter, the \textit{f-measure} as a relation that trades off precision versus recall, is suggested. In fact, it is the weighted harmonic mean of precision and recall, in which both of them are weighted equally:
\begin{equation}
\scriptsize
\label{eq: Equation 4}
f-measure =  \frac{2 \times precision \times recall}{precision + recall}
\end{equation}
Another intuitive measures that simply evaluates the system performance is \textit{accuracy}. As Equation \ref{eq: Equation 5} reveals, it is the proportion of the system assigned (predicted) correct labels outside the total possible observations. \textit{TN}, a factor which is ignored in f-measure, is paid attention here.
\begin{equation}
\scriptsize
\label{eq: Equation 5}
Accuracy =  \frac{TP+TN}{TP+TN+FP+FN}
\end{equation}

\subsection{System Outputs}
Comparing precision and recall reveals that, \textit{FP} and \textit{FN} are two key factors that influence their values. Hence, it is conceivable that they would be considered where the \textit{FP} and \textit{FN} are more important. While, \textit{TP} and \textit{TN} are key factors determining the value of Accuracy. Regarding the APP systems’ functionality, accuracy is preferable to precision and recall, and thereby f-measure. Despite the fact, both of the accuracy and f-measure have been used by researchers for APP systems evaluation. Of course, keeping in mind the ideas behind precision and recall, they would have meaningful interpretations. From that perspective, we decided to achieve them and consequently the f-measure for each of the suggested methods, however that we will basically rely on accuracy.

For the purpose of APP, the predictions were carried out in two steps: APP through five distinct base methods, and APP through ensembling the base methods. The results in each of the two steps are described below. \hyperref[tab: Table 5]{Table~\ref*{tab: Table 5}} provides the results obtained from the four evaluation measures (namely, precision, recall, f-measure, and accuracy) in each of the six suggested APP methods. The performance of each method is evaluated separately in each of the personality traits in the Big Five model (i. e. \{O, C, E, A, N\}).

%(((((((((((((((( Base Methods))))))))))))))))
\subsubsection{Base Methods Outputs}
This paper proposed five distinct methods for APP, each of which performs the prediction independently, during the first step. Specifically, the term frequency vector-based, ontology-based, enriched ontology-based, LSA-based, and deep learning-based (BiLSTM) methods were called into action. 

Among the five proposed base methods, term frequency vector-based method was the first one that unexpectedly obtained the best results for accuracy in all of the five traits, as it can be seen from the \hyperref[tab: Table 5]{Table~\ref*{tab: Table 5}}. This method has also achieved the highest average accuracy value in five traits.

The ontology-based method as well as enriched ontology-based method, approximately have achieved the same results in accuracy for all of the five traits. However that ontology-based method achieved better average accuracy value, enriched ontology-based method slightly acts better in all of the five personality traits, except openness. Nevertheless, insignificant differences between corresponding accuracy values, reveal that enriching the method does not cause to improve the predictions, remarkably. Even though it has worsened the predictions in openness.

Interestingly, the accuracy values for LSA-based method equaled exactly to ontology-based method’s values, except neuroticism, in which the ontology-based method outperformed. Comparing average accuracy values, with a slight difference it has achieved the lowest value.

Deep learning-based methods, generally outperform other methods in most tasks. However, contrary to expectations, there were no significant enhancements in accuracy values, with regard to ontology and LSA-based methods. At the same time, it has finished most predictions (except conscientiousness) in second place. Actually, just the simplest suggested method, namely the term frequency vector-based method, had better results than suggested deep learning-based method. Nevertheless, among the five suggested base methods, the deep learning-based method has achieved the highest precisions for three personality traits and the remained highest two precisions were achieved by the term frequency vector-based method. 

In general, as can be seen in \hyperref[tab: Table 5]{Table~\ref*{tab: Table 5}}, among five suggested base methods for APP, the term frequency vector-based method has achieved best accuracy values in all of the five personality traits in the Big Five model, when they were used independently.

\begin{table}%[htbf]
	\centering
	\caption{Evaluation results for suggested APP methods, including 5 base methods (i.e. TF vector-based, ontology-based, enriched ontology-based, LSA-based, and deep learning-based methods) along with ensemble modeling method}
	\label{tab: Table 5}
	\scriptsize
	\begin{adjustbox}{max width=\textwidth}
		\begin{tabular}{|c|c|ccccc|c|} 
			\hline
			\rowcolor{gray!35} 		%-----> To color the row (not too gray)
			\textbf{Measure} & \textbf{Method} & \textbf{O} & \textbf{C} & \textbf{E} & \textbf{A} & \textbf{N} & \textbf{Avg.}\\ 
			\hline
			\hline
			\multirow{6}*{\textbf{Precision}}
			& TF Vector-Based & 59.84 & 54.87 &	56.95 & 58.04 & 54.92 & 56.92 \\ %\cline{2-8}
			& \cellcolor{gray!15} Ontology-Based  & \cellcolor{gray!15} 51.52 & \cellcolor{gray!15} 50.79 & \cellcolor{gray!15} 51.72 & \cellcolor{gray!15} 53.10 & \cellcolor{gray!15} 49.98 & \cellcolor{gray!15} 51.42 \\ %\cline{2-8}
			& Enriched Ontology-Based & 52.31 & 51.18 & 51.76 & 53.67 & 50.02 & 51.79 \\ %\cline{2-8}
			& \cellcolor{gray!15} LSA-Based & \cellcolor{gray!15} 51.52 & \cellcolor{gray!15} 50.79 & \cellcolor{gray!15} 51.72 & \cellcolor{gray!15} 53.10 & \cellcolor{gray!15} 42.31 & \cellcolor{gray!15} 49.89 \\ %\cline{2-8}
			& Deep Learning-Based (BiLSTM) & \textbf{61.24} & 48.64 & 59.45 & \textbf{69.18} & 50.74 & 57.85 \\ %\cline{2-8}
			& \cellcolor{gray!15} Ensemble Modeling & \cellcolor{gray!15} 58.23 & \cellcolor{gray!15} \textbf{59.32} & \cellcolor{gray!15} \textbf{63.45} & \cellcolor{gray!15} 60.18 & \cellcolor{gray!15} \textbf{61.24} & \cellcolor{gray!15} \textbf{60.48}\\ 
			\hline
			%\hline
			\hline
			
			\multirow{6}*{\textbf{Recall}}
			& TF Vector-Based & 52.40 & 60.65 & 56.50 & 58.17 & 58.88 & 57.32 \\ %\cline{2-8}
			& \cellcolor{gray!15} Ontology-Based  & \cellcolor{gray!15} \textbf{100} & \cellcolor{gray!15} \textbf{100} & \cellcolor{gray!15} \textbf{100} & \cellcolor{gray!15} \textbf{100} & \cellcolor{gray!15} \textbf{100} & \cellcolor{gray!15} \textbf{100} \\ %\cline{2-8}
			& Enriched Ontology-Based & 40.05 & 91.70 & \textbf{100} & 87.63 & \textbf{100} & 83.88 \\ %\cline{2-8}
			& \cellcolor{gray!15} LSA-Based & \cellcolor{gray!15} \textbf{100} & \cellcolor{gray!15} \textbf{100} & \cellcolor{gray!15} \textbf{100} & \cellcolor{gray!15} \textbf{100} & \cellcolor{gray!15} 8.92 & \cellcolor{gray!15} 81.78 \\ %\cline{2-8}
			& Deep Learning-Based (BiLSTM) & 55.22 & 49.38 & 55.84 & 55.65 & 51.48 & 53.51 \\ %\cline{2-8}
			& \cellcolor{gray!15} Ensemble Modeling & \cellcolor{gray!15} 56.54 & \cellcolor{gray!15} 60.16 & \cellcolor{gray!15} 68.32 & \cellcolor{gray!15} 63.12 & \cellcolor{gray!15} 60.15 & \cellcolor{gray!15} 61.66\\ 
			\hline
			%\hline
			\hline
			
			\multirow{6}*{\textbf{F-measure}}
			& TF Vector-Based & 55.87 & 57.62 & 56.73 & 58.10 & 56.83 & 57.03 \\ %\cline{2-8}
			& \cellcolor{gray!15} Ontology-Based  & \cellcolor{gray!15} \textbf{68.00} & \cellcolor{gray!15} \textbf{67.37} & \cellcolor{gray!15} \textbf{68.18} & \cellcolor{gray!15} \textbf{69.37} & \cellcolor{gray!15} 66.65 & \cellcolor{gray!15} \textbf{67.91} \\ %\cline{2-8}
			& Enriched Ontology-Based & 45.37 & 65.69 & 68.22 & 66.57 & \textbf{66.68} & 62.51 \\ %\cline{2-8}
			& \cellcolor{gray!15} LSA-Based & \cellcolor{gray!15} \textbf{68.00} & \cellcolor{gray!15} \textbf{67.37} & \cellcolor{gray!15} \textbf{68.18} & \cellcolor{gray!15} \textbf{69.37} & \cellcolor{gray!15} 14.73 & \cellcolor{gray!15} 57.53 \\ %\cline{2-8}
			& Deep Learning-Based (BiLSTM) & 58.07 & 49.01 & 57.59 & 61.68 & 51.10 & 55.49 \\ %\cline{2-8}
			& \cellcolor{gray!15} Ensemble Modeling & \cellcolor{gray!15} 57.37 & \cellcolor{gray!15} 59.74 & \cellcolor{gray!15} 65.80 & \cellcolor{gray!15} 61.62 & \cellcolor{gray!15} 60.69 & \cellcolor{gray!15} 61.04\\ 
			\hline
			%\hline
			\hline
			
			\multirow{6}*{\textbf{Accuracy}}
			& TF Vector-Based & \textbf{57.36} & 54.68 & 55.41 & 55.45 & 55.29 & 55.64 \\ %\cline{2-8}
			& \cellcolor{gray!15} Ontology-Based  & \cellcolor{gray!15} 51.52 & \cellcolor{gray!15} 50.79 & \cellcolor{gray!15} 51.72 & \cellcolor{gray!15} 53.10 & \cellcolor{gray!15} 49.98 & \cellcolor{gray!15} 51.42 \\ %\cline{2-8}
			& Enriched Ontology-Based & 50.30 & 51.36 & 51.80 & 53.26 & 50.06 & 51.36 \\ %\cline{2-8}
			& \cellcolor{gray!15} LSA-Based & \cellcolor{gray!15} 51.52 & \cellcolor{gray!15} 50.79 & \cellcolor{gray!15} 51.72 & \cellcolor{gray!15} 53.10 & \cellcolor{gray!15} 48.40 & \cellcolor{gray!15} 51.11 \\ %\cline{2-8}
			& Deep Learning-Based (BiLSTM) & 52.83 & 49.39 & 53.85 & 54.86 & 52.43 & 52.67 \\ %\cline{2-8}
			& \cellcolor{gray!15} Ensemble Modeling & \cellcolor{gray!15} 56.30 & \cellcolor{gray!15} \textbf{59.18} & \cellcolor{gray!15} \textbf{64.25} & \cellcolor{gray!15} \textbf{60.31} & \cellcolor{gray!15} \textbf{61.14} & \cellcolor{gray!15} \textbf{60.24}\\ 
			\hline
			%\hline

		\end{tabular}
	\end{adjustbox}
\end{table}

\begin{figure}[htbp]
	
	\scriptsize
	\makebox[\textwidth]{		%fitting figure to textwidth
		\fbox{
			\begin{tikzpicture}
			\centering
			\begin{axis}[
			ybar, axis on top,
			height=8cm, width=12cm,
			bar width=0.225cm,
			ymajorgrids, tick align=inside,
			major grid style={draw=white},
			enlarge y limits={value=.1,upper},
			ymin=0, ymax=80,
			axis x line*=bottom,
			axis y line*=left,
			%axis y line*=right,
			%y label style={at={(0.1,0.5)}},   %decreasing the ylabel distance
			y axis line style={opacity=100},
			tickwidth=0pt,
			enlarge x limits=true,
			legend style={
				draw=none,   %<---- disappear the legend box. delet to show legend box
				at={(0.5,-0.1)},
				legend cell align=left,
				%legend pos= north east,
				anchor=north,
				legend columns=2,	% if = -1 then horizontal legend and = 1 vertical legend
				/tikz/every even column/.append style={column sep=0.5cm}
			},
			ylabel={\textbf{Accuracy} (\%)},
			symbolic x coords={
				Openness,Conscientiousness,Extroversion,Agreeableness,
				Neuroticism},
			xtick=data,
			nodes near coords={
				\pgfmathprintnumber[precision=0]{\pgfplotspointmeta}
			}
			]
			%draw=none will cause to clear the bar borders
			% Different Patterns: horizontal lines, vertical lines, grid, dots, north east lines
			
			\addplot [black,fill=white, postaction={pattern=north west lines}] coordinates {	%----> Term Frequency Vector based Method
				(Openness,57.36)
				(Conscientiousness, 54.68) 
				(Extroversion,55.41)
				(Agreeableness,55.45) 
				(Neuroticism,55.29) };
			\addplot [black,fill=gray] coordinates {	%----> Ontology Based Method
				(Openness,51.52)
				(Conscientiousness, 50.79) 
				(Extroversion,51.72)
				(Agreeableness,53.10) 
				(Neuroticism,49.98)};
			\addplot [black, fill=white] coordinates {	%----> Enriched Ontology Based Method
				(Openness,50.30)
				(Conscientiousness, 51.36) 
				(Extroversion,51.80)
				(Agreeableness,53.26) 
				(Neuroticism,50.06)};
			\addplot [black, fill=gray!30!white , postaction={pattern=horizontal lines}] coordinates {	%----> LSA Based Method
				(Openness,51.52)
				(Conscientiousness, 50.79) 
				(Extroversion,51.72)
				(Agreeableness,53.10) 
				(Neuroticism,48.40)};
			\addplot [black,fill=white, postaction={pattern=north east lines}] coordinates {	%----> Deep Learning Based Method		 addplot [draw=none, fill=red!50] coordinates
				(Openness,52.83)
				(Conscientiousness, 49.39) 
				(Extroversion,53.85)
				(Agreeableness,54.86) 
				(Neuroticism,52.43)};
			
			\addplot [black,fill=white, postaction={pattern=dots}] coordinates {	%----> Ensemble Learning		 addplot [draw=none, fill=red!50] coordinates
				(Openness,56.30)
				(Conscientiousness, 59.18) 
				(Extroversion,64.25)
				(Agreeableness,60.31) 
				(Neuroticism,61.14)};
			
			\legend{Term Freq. Vector-Based Method,Ontology-Based Method,Enriched Ontology-Based Method, LSA-Based Method, Deep Learning-Based Method, Ensemble Modeling}
			
			\end{axis}
			\end{tikzpicture}
		}
	}
	\caption{\protect\raggedright Accuracy values (as the most appropriate evaluation metric in APP) for suggested APP methods, including 5 base methods along with ensemble modeling method in each of the five personality traits in Big Five model (results are rounded)}
	
	\label{fig: Figure 3}		%---- fig 2
\end{figure}
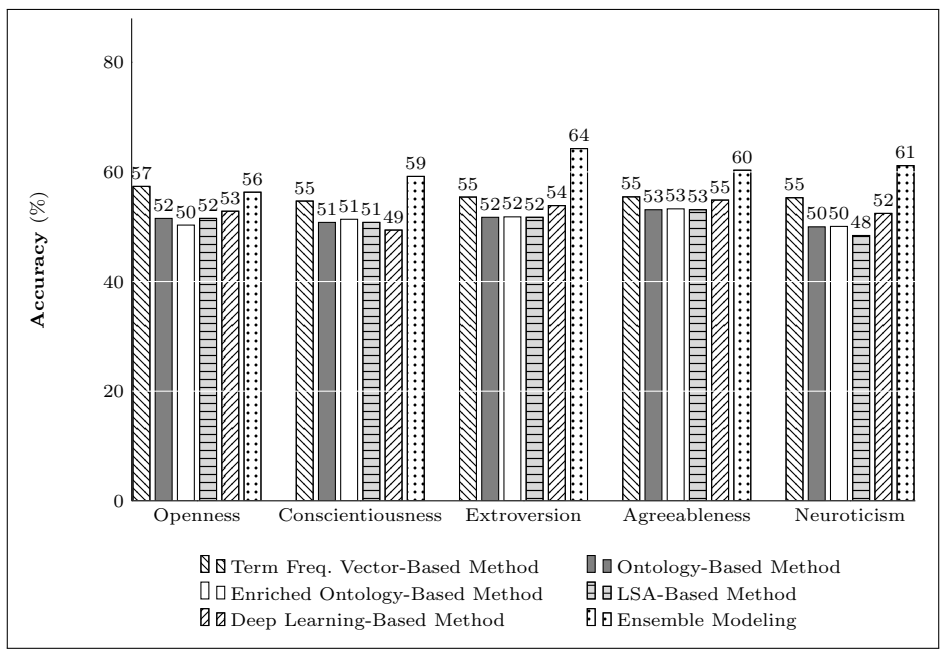

%(((((((((((((((( Ensemble Modeling))))))))))))))))
\subsubsection{Ensemble Modeling Outputs}
In the second step toward ensemble modeling, we used a HAN as the meta-model to implement the stacking algorithm. Comparing average accuracy values among six suggested APP methods (five base methods and an ensemble modeling method), ensemble modeling method achieved the highest value, as can be seen from \hyperref[tab: Table 5]{Table~\ref*{tab: Table 5}}. It also outperforms in all of the five personality traits, except openness in which the term frequency vector-based method slightly acts better. Moreover, this method has achieved the highest average precision among all methods. Consequently, as hypothesized our experiments prove that the prediction error decreases when the ensemble method is used.

\hyperref[fig: Figure 3]{Figure~\ref*{fig: Figure 3}}, compares the accuracy of all suggested methods (five base method as well as ensemble modeling method), in predicting openness, conscientiousness, extroversion, agreeableness, and neuroticism over the Essays Dataset.

%%%%%%%%%%%%%%%%%%%%%%%%%%%%%%%%%%%%%%%%%%%%%%%%%%%%%%%%%%%%%%%%%%%%%%%%%%%%%%%%%%%%%%%%%%%%%%%%%%%%%%%%%%%%%%%%%
\section{Discussion}
Prior to discussing the results of suggested methods for APP, particular attention must be paid to some facts about personality prediction. Indeed, personality prediction is really a complicated task. Actually, it is correct even for human. One cannot deny that people face a challenge even when they are analyzing and expressing their own personality. What is more, people usually act conservatively when writing and speaking. Besides, the formal style of writing or speech undoubtedly will conceal some facts about people personality. Over and above, as previously stated, each of the five personality traits, represents a range between two extremes. Obviously, the binary classification of a continuum will cause to miss some information about personality traits. Respecting the facts, it should be confessed that the personality prediction task is really complicated both for human and for machine.

As it can be inferred from \hyperref[tab: Table 5]{Table~\ref*{tab: Table 5}}, among the five suggested base methods, unexpectedly the term frequency vector-based method in spite of its simplicity, generally outperforms all other base methods when they are applied independently. Taking into account the simplicity of the method and the complexity of the task, \textit{Occam’s Razor} springs to one's mind; it is most likely that, the simplest solution is the right one. At the same time, the deep learning-based method that was expected to achieve the highest accuracies, has achieved the second best accuracies. This would be as a result of suggested deep network’s less ability of prediction, due to the less amounts of personality related informational content in essays. Notably, this method achieves the highest precision in most (three from five) personality traits. It means that, in spite of low accuracies, deep learning-based method causes the most precise predictions. Of course, we believe that considering the task’s complexity, enriching the dataset will cause to improve the accuracies for deep learning methods (however it will also improve other method’s accuracies). It must be said that, we have also designed an LSTM network for APP, but it could not overcome the BiLSTM’s outcomes. 

Among the five suggested base methods the three remained methods approximately perform equally. Using ontology lexical database and its semantic networks among words, results same predictions to LSA-based method. The Ontology-based method’s poor results approve that, the existed semantic network among words cannot be regarded as a distinguishing criterion for personality traits. Even enriching the method with the most correlated words in five personality traits, does not improve the results remarkably. We believe that modeling the vast semantic network among words, potentially increases the \textit{oversensitivity} of the model. The resulted values for recall and precision for this methods, support this idea; approximately all of the expected items (100\% for ontology-based and about 84\% for enriched method) are predicted correctly (recall), but just about half of the predictions are labeled correctly (precision). In addition, a possible explanation for this outcome is that, ontologies formally specify the \textit{types} rather than \textit{type samples}. 

Likely, LSA-based method is affected by oversensitivity, that despite its ability in latent semantic relation acquisition among words, causes to poor accuracies. Moreover, comparing the obtained results from LSA and term frequency-based method points to the fact that, term frequency vectors (in which each dimension is dedicated to a unique word) are more successful than topic vectors (in which each dimension is dedicated to a unique extracted topic). It can be interpreted as a justification of the lexical hypothesis; more important personality characteristics are more likely to be encoded in a single word \cite{2018wiley}.

Generally, results approve that among five suggested base methods the simple term frequency-based method is so competent to predict personality in all five personality traits in the Big Five model, albeit it loses the relative ordering of the terms in essays.

At last, we finally achieved the major objective of our study; we enhanced the accuracy of APP using ensemble modeling. Actually, it is due to the substantial ability of stacking. During training, the meta-model was trained using both Essays Dataset’s essays as well as their actual labels (gold standard), and base methods’ predictions (ground truth). Practically, it compares the base methods’ predictions with actual corresponding labels for each essay. Afterwards, during final classification it assigns more weights to those methods with correct classifications. Consequently, the accuracy of APP was improved.

\hyperref[tab: Table 6]{Table~\ref*{tab: Table 6}} compares the accuracy and F-measure values of our proposed method and several state-of-the-art APP methods whose hypotheses are approximately the same as ours. Specifically, those researches that are concentrated on automatic personality prediction from text, in Big Five model using Essays Dataset. Generally, as highlighted in \hyperref[tab: Table 6]{Table~\ref*{tab: Table 6}}, our proposed APP method (namely, ensemble modeling) has achieved the highest values both for accuracy and F-measure averages amongst others. To be more accurate, it has achieved the highest accuracies in all of the personality traits, except openness.

\begin{table}%[htbf]
	\centering
	\caption{Comparison between our proposed method's results and state of the art researches in APP from text
	\newline \footnotesize (there is no information available for absent values)}
	\label{tab: Table 6}
	\scriptsize
	\begin{adjustbox}{max width=\textwidth}
		\begin{tabular}{|c|c c c c c | c||c c c c c | c|}
			\hline
			
			\multirow{2}{2.5cm}{\centering \textbf{APP Method}}& \multicolumn{6}{p{7cm}||}{\centering \cellcolor{yellow!15} \textbf{F-measure}} & \multicolumn{6}{p{7cm}|}{\centering \cellcolor{yellow!15} \textbf{Accuracy}}\\
			\cline{2-13} & \multicolumn{1}{c}{\textbf{O}} & \multicolumn{1}{c}{\textbf{C}} & \multicolumn{1}{c}{\textbf{E}} & \multicolumn{1}{c}{\textbf{A}} & \multicolumn{1}{c|}{\textbf{N}} & \multicolumn{1}{c||}{\textbf{Avg.}} & \multicolumn{1}{c}{\textbf{O}} & \multicolumn{1}{c}{\textbf{C}} & \multicolumn{1}{c}{\textbf{E}} & \multicolumn{1}{c}{\textbf{A}} & \multicolumn{1}{c|}{\textbf{N}} & \multicolumn{1}{c|}{\textbf{Avg.}}\\  
			\hline
			\hline
			Our Proposed Method & 57.37 & \textbf{59.74} & 65.80  & \textbf{61.62} & \textbf{60.69} & \textbf{61.04} & 56.30 & \textbf{59.18}  & \textbf{64.25} & \textbf{60.31} & \textbf{61.14} & \textbf{60.24}\\
			\rowcolor{gray!25}
			
			Tighe et al. \cite{Tighe2016PersonalityTC} & 61.90 & 56.00 & 55.60  & 55.70 & 58.30 & 57.50 & 61.95 & 56.04  & 55.75 & 57.54 & 58.31 & 57.92\\

			Salminen et al. \cite{Salminen2020Joni} & \textbf{65.30} & 54.30 & \textbf{66.20}  & 60.30 & 33.20 & 55.86 &  &   &  &  &  & \\
			\rowcolor{gray!25}
			
			Majumder et al. \cite{7887639} &  &  &   &  &  &  & \textbf{62.68} & 57.30  & 58.09 & 56.71 & 59.38 & 58.83\\

			Kazameini et al. \cite{kazameini2020personality} &  &  &   &  &  &  & 62.09 & 57.84  & 59.30 & 56.52 & 59.39 & 59.03\\
			\rowcolor{gray!25}
			\hline
		\end{tabular}

	\end{adjustbox}
\end{table}

%%%%%%%%%%%%%%%%%%%%%%%%%%%%%%%%%%%%%%%%%%%%%%%%%%%%%%%%%%%%%%%%%%%%%%%%%%%%%%%%%%%%%%%%%%%%%%%%%%%%%%%%%%%%
\section{Conclusion and Future Works}
Nowadays, texts provide a large amount of human interactions in different information infrastructures (social media, email, etc.). Automatic Personality Prediction (APP) from the text, uncovers the personality characteristics of the writers, without facing or even knowing them. Enhancing the accuracy of APP, will step researches ahead toward automating the analysis of the relation between personality and various human behaviors (rather than pure APP). For that purpose, in this study we have presented an ensemble modeling method that was based on five new APP methods, including term frequency vector-based, ontology-based, enriched ontology-based, latent semantic analysis-based, and deep learning-based (BiLSTM) methods. We have obtained satisfactory results proving that as we have hypothesized, ensembling the base methods enhances the accuracy of APP. Moreover, we have found that when the five APP methods have been used independently, the simple term frequency vector-based method outperforms all others. 

Each of the five personality traits represents a range between two extremes. This means that the Big Five model, represents the personality of individuals in a five dimensional space. In simple words, contrary to binary classification, each of the five personality traits may have different values, indicating the intensity of each trait. Thereby, fuzzy classification sounds so much better than binary classification. Further works are required to establish this idea. 

Reaching maturity and meeting acceptable accuracies via computational researches in personality prediction, will attract APP researchers’ attention to focus on automatic analyzing the relationship between personality and different human behaviors, promptly. Namely, automatic prediction of: aggressive and violent behavior, antisocial behavior, delinquency, good citizenship and civic duty, marital instability, political attitude, orienting voting choices, IoT, etc\footnote{A more complete list is provided at the beginning of section 2}. This topics also are deferred to future work.

Moreover, as it can be inferred, prediction or perception is prior to generation in personality analysis, in which its quality is determining factor in posterior. Achieving acceptable performance in APP, will facilitate investigations in \textit{Automatic Personality Generation (APG)}. As a consequence, it is recommended that further researches should be undertaken in APG.

%%%%%%%%%%%%%%%%%%%%%%%%%%%%%%%%%%%%%%%%%%%%%%%%%%%%%%%%%%%%%%%%%%%%%%%%%%%%%%%%%%%%%%%%%%%%%%%%%%%%%%%%%%%%
\section{Acknowledgments}
This project is supported by a research grant of the University of Tabriz (number S/806).

\section{Declarations}
\textbf{Funding} This study was funded by the University of Tabriz (grant number S/806).\\
\textbf{Conflict of Interest} All of the authors declare that they have no conflict of interest against any company or institution.
\\
\textbf{Ethical standards} This article does not contain any studies with human participants or animals performed by any of the authors.
%\bibliographystyle{spbasic-nosort}

\bibliographystyle{unsrt}      % basic style, author-year citations
%\footnotesize
\bibliography{Bibliography}   % name your BibTeX data base

\end{document}